\documentclass{article}

\usepackage{arxiv}

\usepackage[utf8]{inputenc} 
\usepackage[T1]{fontenc}    
\usepackage{hyperref}       
\usepackage{url}            
\usepackage{booktabs}       
\usepackage{amsfonts} 
\usepackage{amssymb}
\usepackage{nicefrac}       
\usepackage{microtype}      
\usepackage{lipsum}		
\usepackage{graphicx}
\usepackage{natbib}
\usepackage{doi}

\usepackage{algorithm2e}
\usepackage{multirow} 
\usepackage{multicol}
\usepackage{subfigure}
\usepackage[flushleft]{threeparttable}
\usepackage{array}
\usepackage{mathtools}
\usepackage{tikz}
 \usepackage{lscape}
\title{ORS: A novel Olive Ridley Survival inspired Meta-heuristic Optimization Algorithm}

\author{ {Niranjan Panigrahi, Sourav Kumar Bhoi, Debasis Mohapatra, Rashmi Ranjan Sahoo}\\
	Department of CSE\\
	 PMEC, Berhampur \\
	India, 761003 \\
	\texttt{\{niranjan.cse, sourav.cse, debasis.cse, rashmiranjan.cse\}@pmec.ac.in} \\
	  \AND
	{Kshira Sagar Sahoo} \\
	Department of Computing Science\\
	Ume\r{a} University, Ume\r{a}\\
	Sweden, 901 87 \\
	\texttt{kshirasagar12@gmail.com} \\
        \And
	{Anil Mohapatra} \\
	Estuarine Biology Regional Center,\\
 Zoological Survey of India,\\Gopalpur-on-Sea - 761002, Ganjam, India\\
	\texttt{anil2k7@gmail.com} \\
}

\renewcommand{\shorttitle}{\textit{arXiv} Template}

\hypersetup{
pdftitle={A template for the arxiv style},
pdfsubject={q-bio.NC, q-bio.QM},
pdfauthor={David S.~Hippocampus, Elias D.~Striatum},
pdfkeywords={First keyword, Second keyword, More},
}

\begin{document}
\maketitle

\begin{abstract}
	Meta-heuristic algorithmic development has been a  thrust area of research since its inception. In this paper, a novel meta-heuristic optimization algorithm, Olive Ridley Survival (ORS), is proposed which is  inspired from survival challenges faced by hatchlings of Olive Ridley sea turtle. A major fact about survival of Olive Ridley reveals that out of one thousand Olive Ridley hatchlings which emerge from nest, only one survive at sea due to various environmental and other factors. This fact acts as the backbone for developing the proposed algorithm. The algorithm has two major phases: hatchlings survival through environmental factors and impact of movement trajectory on its survival.  The phases are mathematically modelled and implemented along with suitable input representation and fitness function. The algorithm is analysed theoretically. To validate the algorithm, fourteen mathematical benchmark functions from standard CEC test suites are evaluated and statistically tested. Also, to study the efficacy of ORS on recent complex benchmark functions, ten benchmark functions of CEC-06-2019 are evaluated. Further, three well-known engineering problems are solved by ORS and compared with other state-of-the-art meta-heuristics. Simulation results show that in many cases, the proposed ORS algorithm outperforms some state-of-the-art meta-heuristic optimization algorithms. The sub-optimal behavior of ORS in some recent benchmark functions is also observed.
\end{abstract}


\section{Introduction}
 Optimization problem in the field of mathematics and real-world domains has driven the development of novel heuristics and meta-heuristics strategies since last few decades \cite{r1}, \cite{evolving1}. The optimization problem may be of constrained or unconstrained type. These strategies are mostly suitable for solving NP-hard problems which lead to near optimal solutions. 
The existing meta-heuristics broadly fall into two categories, population based and local search based. In population based approach, the solution is searched from a set of feasible solutions as input until a near optimal solution is found out. The search process terminates when maximum number of iteration reached or there is no change in optimal value which is already achieved. In the later category, the search starts with a random feasible solution and the search process iterates with small changes in each iteration until a satisfactory solution is obtained. Although a plethora of meta-heuristics are developed since its inception, it is still continuing as an open area to develop novel meta-heuristics.  In this context, an extensive survey has been carried out and summarized below to highlight a set of well-known and recent meta-heuristics.  \par
The nature-inspired meta-heuristic algorithms have demonstrated remarkable performance in solving various hard optimization problems. These algorithms are generally inspired by natural phenomena, social dynamics, physical systems, etc. to solve hard optimization problems through approximation that is achievable by the help of stochastic parameters. In this context, three major classes of meta-heuristic algorithms namely Evolutionary-based, Behaviour-based, and Physics-based algorithms are discussed. This classification is depicted through a hierarchical representation in Figure 1.\par 
\begin{figure}[h]
    \centering
    \includegraphics[width=1\textwidth]{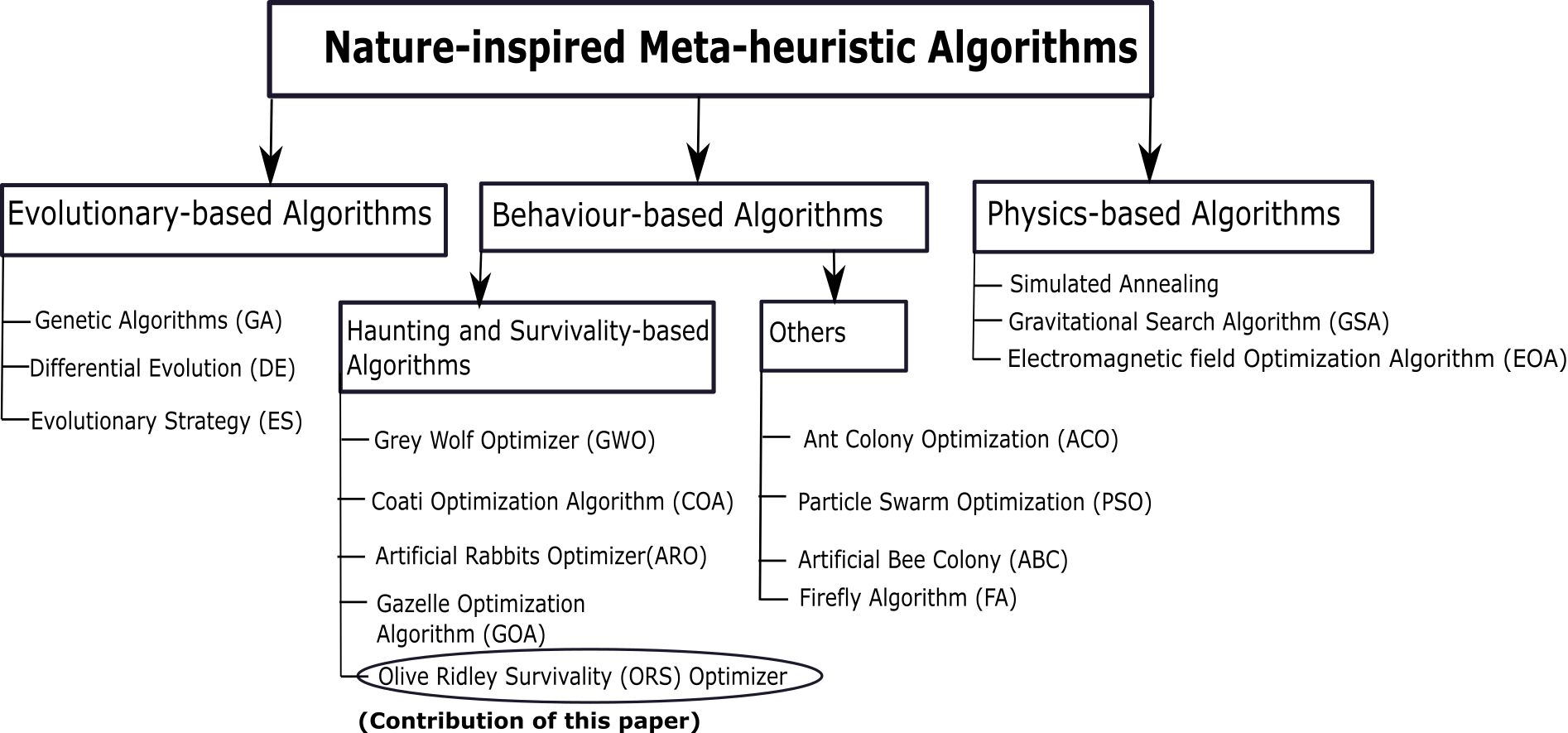}
    \caption{Major classes of nature-inspired meta-heuristic algorithms}
    \label{fig:mesh1}
\end{figure}

The founding stone of meta-heuristic algorithms was laid by the group of algorithms inspired by Darwin's theory of evolution. Holland presented a seminal work named Genetic Algorithms (GA) \cite{29} which is based on Darwin's theory of evolution. Algorithms like Evolutionary Strategy (ES) \cite{30}, Differential Evolution (DE) \cite{1}, etc. adopt the same idea. Some of the algorithms mimic the behaviors of different creatures. For example, Ant Colony Optimization (ACO)\cite{28} follows the foraging behavior of real ants, and Particle Swarm Optimization (PSO) \cite{31} is inspired by the group behavior of birds and fishes. A few other algorithms in this category are Artificial Bee Colony (ABC) Algorithm \cite{32}, Firefly Algorithm (FA)\cite{33}, etc. A separate group of meta-heuristic algorithms are inspired by physical phenomena. Simulated Annealing (SA) \cite{34} is based on the annealing process of metal. The Gravitational Search Algorithm (GSA)\cite{35} is inspired by the law of gravitation and the motion of celestial bodies. Electromagnetic field Optimization Algorithm (EOA) \cite{36} mimics the electromagnetic theory.    

Among various nature-inspired meta-heuristic algorithms, some algorithms are inspired by haunting and survivability behavior. The Cuttlefish Algorithm (CA) \cite{3} uses reflection and visibility to determine the color of the Cuttlefish. The color change helps the Cuttlefish escape from predators and is also useful in hunting for food. Grey Wolf Optimization (GWO)\cite{4} mimics the leadership categories of grey wolves used for searching, encircling, and haunting prey. The whale Optimization Algorithm (WOA)\cite{11} characterizes the bubble net haunting strategy for attacking the prey. Coati Optimization Algorithm (COA) \cite{23} follows the haunting and escape behaviors of Coati. Artificial Rabbits Optimizer(ARO) \cite{26} is based on detour foraging and random hiding behaviors. Gazelle Optimization Algorithm (GOA) \cite{27} imitates the grazing behavior of Gazelle when the predator is not spotted in the exploitation phase and considers the random runaway as the exploration phase once the predator is spotted. Some crucial works in this field are listed in Table 1 for quick reference.

It can be concluded from above highlighted survey that the development of novel meta-heuristics are based on two fundamental principles: exploration or diversification and exploitation or intensification of the underlying search space \cite{peraza2021bio}. The effectiveness of any meta-heuristic depends on how optimally it can maintain a balance between these two processes. All the meta-heuristic algorithms simulate the aspects of exploration and exploitation to meet global optimum. The multiple candidates of such algorithms with their non-localized deployment enable the exploration capacity to cover the search space and exploit the local optimum in parallel at the local levels. Eventually all local optimums with comparison lead to a global optimum solution. 
 
Though a series of meta-heuristic algorithms are developed since last few decades, there is always a scope for development of new and novel meta-heuristics. This is in accordance with the theory of ‘’No Free Lunch (NFL)’’ which says that no single meta-heuristic can solve all optimization problems. This motivates the continuous development of novel meta-heuristics which can improve the performance of different type and dimension of optimization problems. Following this assertion, we have made an attempt to study the survival challenges of Olive Ridley sea turtle which is listed worldwide in the group of endangered species.  A  major fact about survival of Olive Ridley reveals that out of one thousand Olive Ridley hatchlings which emerge from nest, only one survive at sea due to various environmental and other factors. This fact acts as the backbone for developing the proposed algorithm. To this end, 
the major contributions in this paper are as follows.
\begin{itemize}
    \item This work proposes a novel meta-heuristic algorithm, ORS, based on survival phenomenon of hatchlings of Olive Ridley sea turtle.
    \item The work mainly consists of two phases, environmental impact and movement trajectory impact on survival of hatchlings. 
    \item The environment factors considered are sand temperature, emergence order from nest, and time of the day. The movement trajectory is modeled as curvilinear motion. This two phases along with stochastic control parameters facilitate exploitation and exploration. 
    \item An in-depth theoretical analysis is presented for the proposed algorithm.
    \item  Benchmark functions from standard CEC test suite are considered for validating the ORS algorithm and are tested statistically.
    \item Well-known engineering problems are optimally solved by ORS and compared with state-of-the-art meta-heuristics.
\end{itemize}

The rest of the work are discussed as follows. Section 2 presents the background investigation of Olive Ridley. Section 3 presents the modelling part of the ORS algorithm. Section 4 discusses about the ORS optimizer and description of the algorithm in detail. Section 5 shows simulation of the algorithms on standard CEC test suite benchmark functions and comparative analysis with state-of-the-art meta-heuristic and presents the application of ORS on three well-known engineering problems. Finally section 6 concludes and presents the future scope of the work.

\begin{table}
\caption{Some prominent meta-heuristic algorithms}
\scriptsize
\label{tab:1}       
\begin{tabular}{p{5cm}p{1cm}p{6cm}}
\hline\noalign{\smallskip}
Algorithm & Year & Source of Inspiration \\
\noalign{\smallskip}\hline\noalign{\smallskip}

Differential Evolution (DE)\cite{1} & 1996 & Darwin’s theory of evolution \\ 
Bat Algorithm (BA)\cite{2} &2012 & Echolocation behavior of bats \\

Cuttlefish Algorithm (CFA)\cite{3} & 2013&Color changing behavior of cuttlefish\\

Grey Wolf Optimizer (GWO) \cite{4} & 2014 & Hunting strategy of grey wolves using leadership hierarchy\\

Coral Reefs Optimization (CRO) \cite{5} &2014& Fight for space and reproduction strategies of corals  \\

Chicken Swarm Optimization (CSO) \cite{6} &2014& Hierarchical ordering in chicken swarms\\

Bird Swarm Algorithm (BSA) \cite{7}&2015& Social connection structure in bird swarms\\

Monkey Algorithm (MA) and the Krill Herd Algorithm (KHA)\cite{8} &2015 & Two existing algorithms MA and KHA (Hybridization)\\

Muti-Verse Optimizer (MVO) \cite{9} &2015 & Concepts of white holes, black holes, and wormholes\\
Sine Cosine Algorithm (SCA)\cite{10} &2016& Mathematical model using sine cosine functions\\
Whale Optimization Algorithm (WOA) \cite{11} & 2016 & Hunting behavior of humpback whales\\

Salp Swarm Algorithm (SSA) \cite{12}&2017& Navigation and foraging behaviors of salps\\

Spotted Hyena Optimizer (SHO) \cite{13}&2017& Collaboration behaviour of spotted hyenas\\

Butterfly Optimization Algorithm (BOA) \cite{14} & 2018 & Foraging and mating behavior of butterflies\\

Earthworm Optimisation Algorithm (EWA) \cite{15}&2018& Reproduction of earthworms\\

Emperor Penguin Optimizer (EPO) \cite{16}&2018&Huddling behavior of emperor penguins\\

Black Widow Optimization Algorithm (BWO)\cite{17}&2020& Mating behaviour of black widow spider\\

Manta Ray Foraging Optimization (MRFO) \cite{18}&2020& Foraging behavior of manta rays\\

Tunicate Swarm Algorithm (TSA) \cite{19} & 2020& Navigation and foraging behaviors of Tunicates\\

Rat Swarm Optimizer (RSO) \cite{20}&2020& Chasing and attacking behaviors of rat\\

Artificial Hummingbird Algorithm (AHA)\cite{21} & 2021 & Flight skills and foraging behaviour of hummingbirds\\

 Cat and Mouse-Based Optimizer (CMBO)\cite{22}&2021& Natural behavior between cats and mice\\
 
 Coati Optimization Algorithm (COA)\cite{23}&2022 & Hunting and escape strategies of coati\\

Tasmanian Devil Optimization (TDO)\cite{24} & 2022 & Feeding behavior of  Tasmanian devils\\

Zebra Optimization Algorithm (ZOA)\cite{25}&2022& Foraging and defense strategies 0f Zebras\\

\noalign{\smallskip}\hline
\end{tabular}
\end{table}


\section{Background Investigation}
This section briefly describes the Olive Ridley life cycle and its behaviour. Further, a summary of its adapted strategies are presented to show its survival phenomena and challenges faced during its birth phase to survive which acts as the motivation for this research. 
\begin{figure}[h]
    \centering
    \includegraphics[width=6cm]{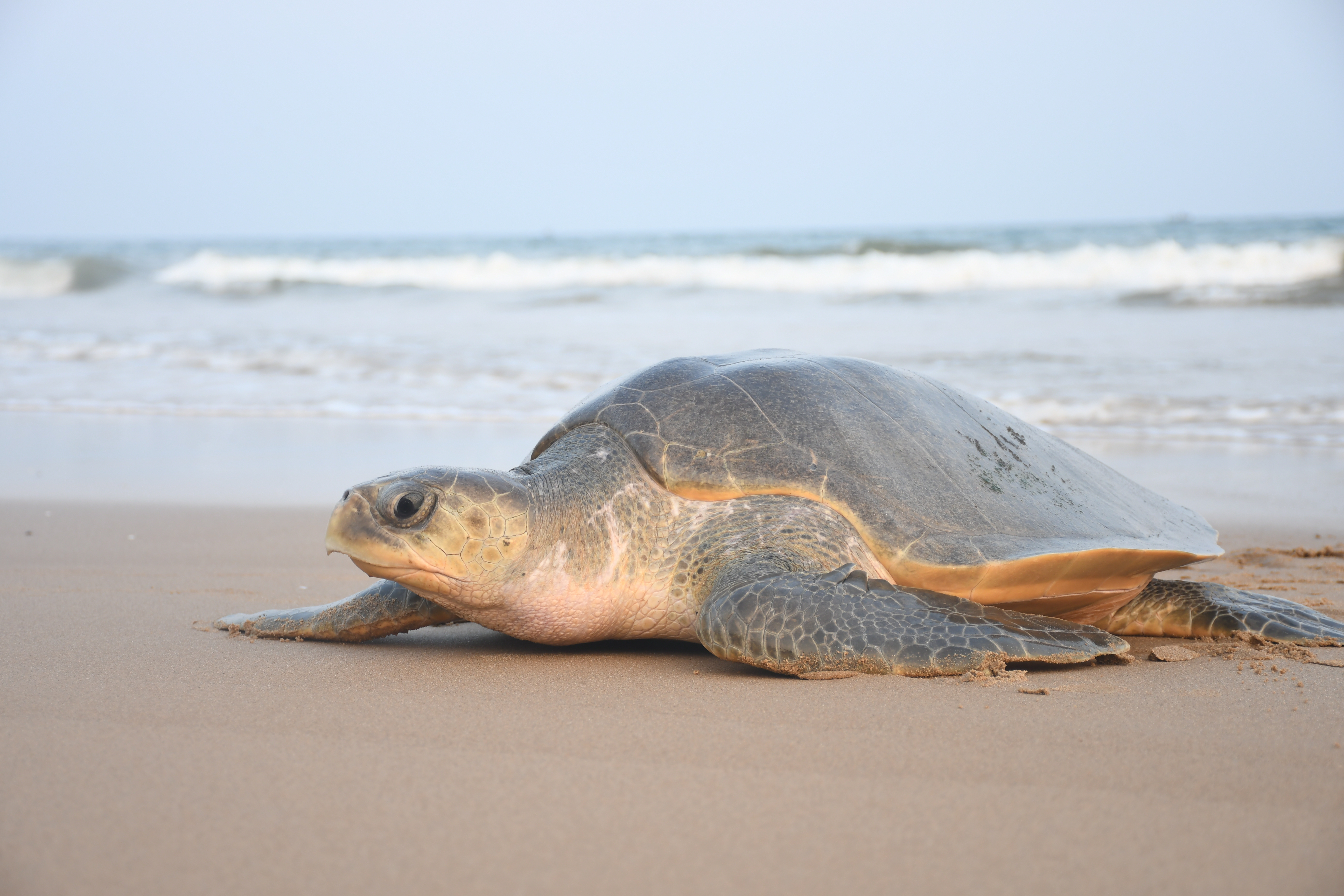}
    \caption{Olive Ridley picture taken by Estuarine Biology Regional Center, Zoological Survey of India, Gopalpur-on-Sea, Ganjam, India}
    \label{fig:galaxy}
\end{figure}
\subsection{Olive Ridley LIfe Cycle \& Behaviour}
The Olive Ridley sea turtle, scientifically known as Lepidochelys olivacea as shown in Fig. 2, is both the most numerous and diminutive of the seven kinds of turtles found on Earth. Because of the direct effect they have on other marine life,  sea turtles are thought to be essential to the health of the ocean.  Thus, turtles are generally thought of as an integral component of marine ecosystems \cite{r2}. Olive ridleys are known for their "arribadas" (Spanish for "arrivals"), which are mass nesting events that only occur in a few unique topography locations of Odisha's Rusikulya, Gahiramatha, and chandrabhaga in India, Costa Rica, and Mexico. Olive ridleys have a worldwide distribution that encompasses the entire circumtropical region. In the course of this spectacular event, thousands of female turtles congregate at the same time on particular nesting beaches in order to lay their eggs. This coordinated nesting behavior produces a captivating show along the shoreline, usually at night. The turtles painstakingly excavate their nests in the fine sand, laying a profusion of eggs and then sanding them down.

The migratory behavior exhibited by numerous Olive Ridley turtle populations is an effective biological strategy for efficiently locating prey that is unevenly distributed. Olive ridleys are the most common sea turtles, which may be explained by their adaptability to the sudden changes in dynamic habitats. Nevertheless, sea turtle numbers worldwide have been steadily decreasing for several decades. As a result of this significant decrease, all turtle species have been categorized as vulnerable, endangered, or critically endangered within the marine ecosystem. Olive ridleys encounter several hazards while they are near the sea. Thus, the loss of natural nesting habitats for olive ridleys is mostly caused by casuarina plantations, beach erosion, the development of tourist complexes, artificial lighting, and egg and hatchling predation. Olive ridleys are vulnerable to a number of hazards in oceanic environments, including pollution, harmful algal blooms, directed fishing net capture, bycatch, and marine trash, all of which have the potential to be lethal.

\subsection{Survival Phenomena \& Motivation}
This section presents the natural survival strategies and motivation behind this research.
\subsubsection{Survival Phenomena}
The Olive Ridley sea turtle has evolved a holistic strategy to survival by utilizing a combination of reproductive, behavioral, and physiological tactics. This demonstrates its impressive flexibility to the obstacles posed by both terrestrial and marine habitats.\\

\textit{Mass Nesting (Arribada)}: One of the unique survival tactics employed by Olive Ridleys is the phenomena known as "Arribada," or mass nesting. The coordinated nesting habit of hundreds of female turtles entails their simultaneous convergence on particular beaches to deposit their eggs. By overwhelming predators, this tactic raises the likelihood that some eggs will hatch. \textit{Egg Placement and Camouflage}: The female Olive Ridley meticulously chooses and excavates nests in the sandy beaches, delicately places their eggs and cover these with sand to disguise them. The choice of strategically hidden nesting locations aids in shielding the eggs from potential predators and environmental hazards.\\

\textit{Temperature-Dependent Sex Determination}: The Olive Ridley sea turtle exhibits temperature-dependent sex determination. The sex of the hatchlings is influenced by the ambient temperature during incubation; warmer temperatures produce more females. The species is more resilient to changes in its environment because of its adaptation. \textit{Habitat Selection}: Olive Ridley turtles demonstrate a discerning choice of environments for the purposes of foraging, mating, and nesting. The selection of these options is determined by various circumstances, including water temperature, prey abundance, and appropriate nesting locations, which collectively contribute to the species' overall survival and reproductive achievements.

\subsubsection{Motivation}
Despite of several natural survival tactics adapted  by this species, during birth phase, Olive Ridley turtles known as hatchlings,  strive to survive due to various environmental and other indirect factors. One alarming observation is that out of one thousand Olive Ridley hatchlings which emerge from nest, only one survive at sea \cite{r2}. Once, hatchlings emerge from their nests, they start crawling towards sea as shown in Fig. 3. During their movement towards sea, various environmental factors, e.g., sand temperature, emergence order from nests, and time of the day of the crawling affect their speed of movement \cite{r2}. They suffer thermal stress during high temperature period of the day which causes death of some hatchlings.
\par Hatchlings emergence order from nests is also a major factor of their momentum towards sea which affects their survival. The hatchlings emerging early crawl faster than hatchlings emerging late. In case of slower speed, the hatchlings have high chance of exposure to land predators and avian. Another indirect factor that affects locomotion of hatchlings is their movement trajectory. During their movement on sea shore, they face debris and as a result, they change their direction of movement. If they deviate from straight line movement, time to reach to sea increases and  they are supposed to expose towards predators.  In summary, hatchlings' survival depends on how fast they can reach to sea. But, their survival is affected by their crawling speed and in turn, affect their fitness.
This motivates the development of the proposed meta-heuristic.
\begin{figure}[h]
    \centering
    \includegraphics[width=8cm]{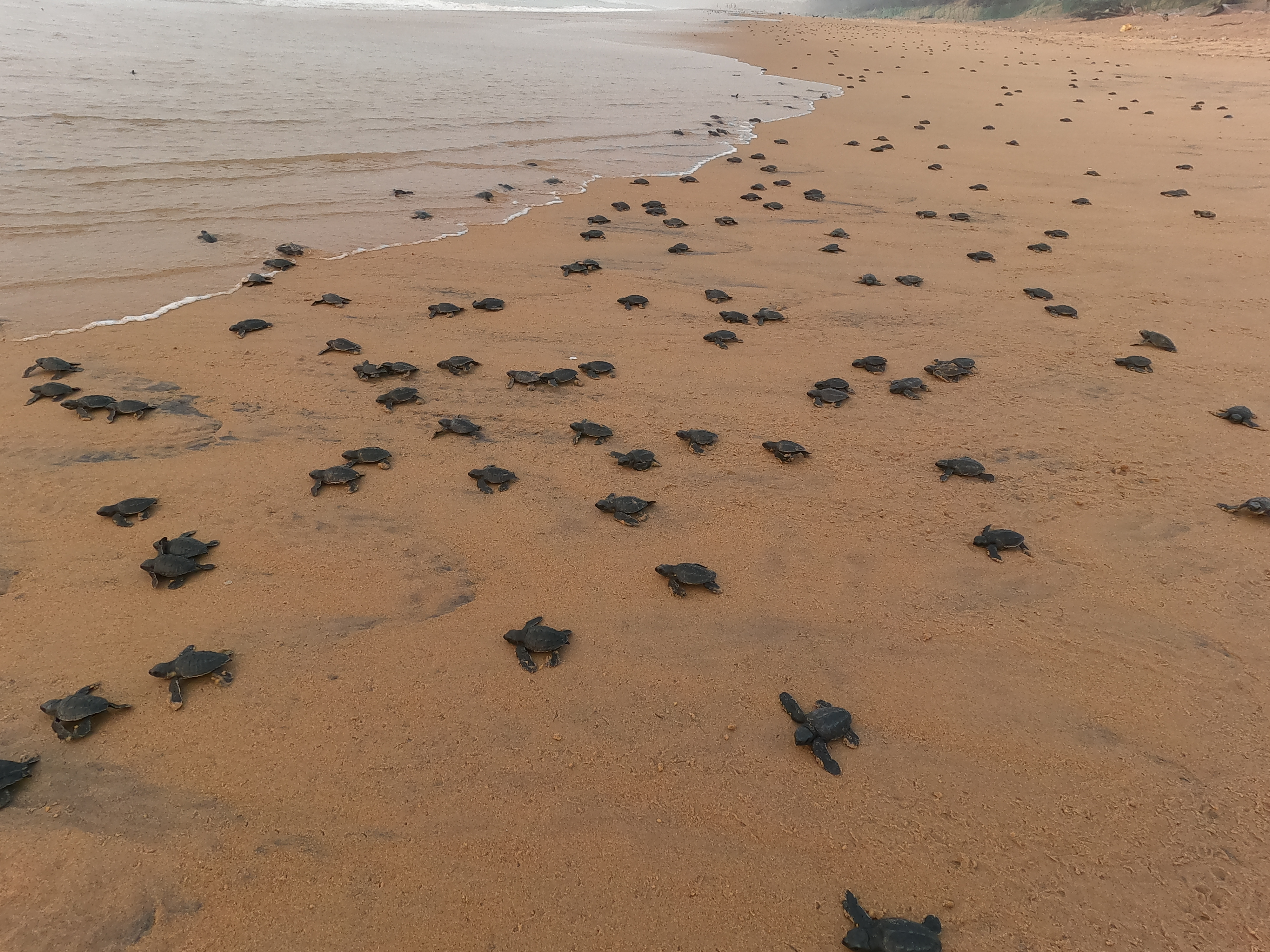}
    \caption{Hatchlings moving towards sea}
    \label{fig:galaxy}
\end{figure}
\begin{figure}[h]
    \centering
    \includegraphics[scale=0.7]{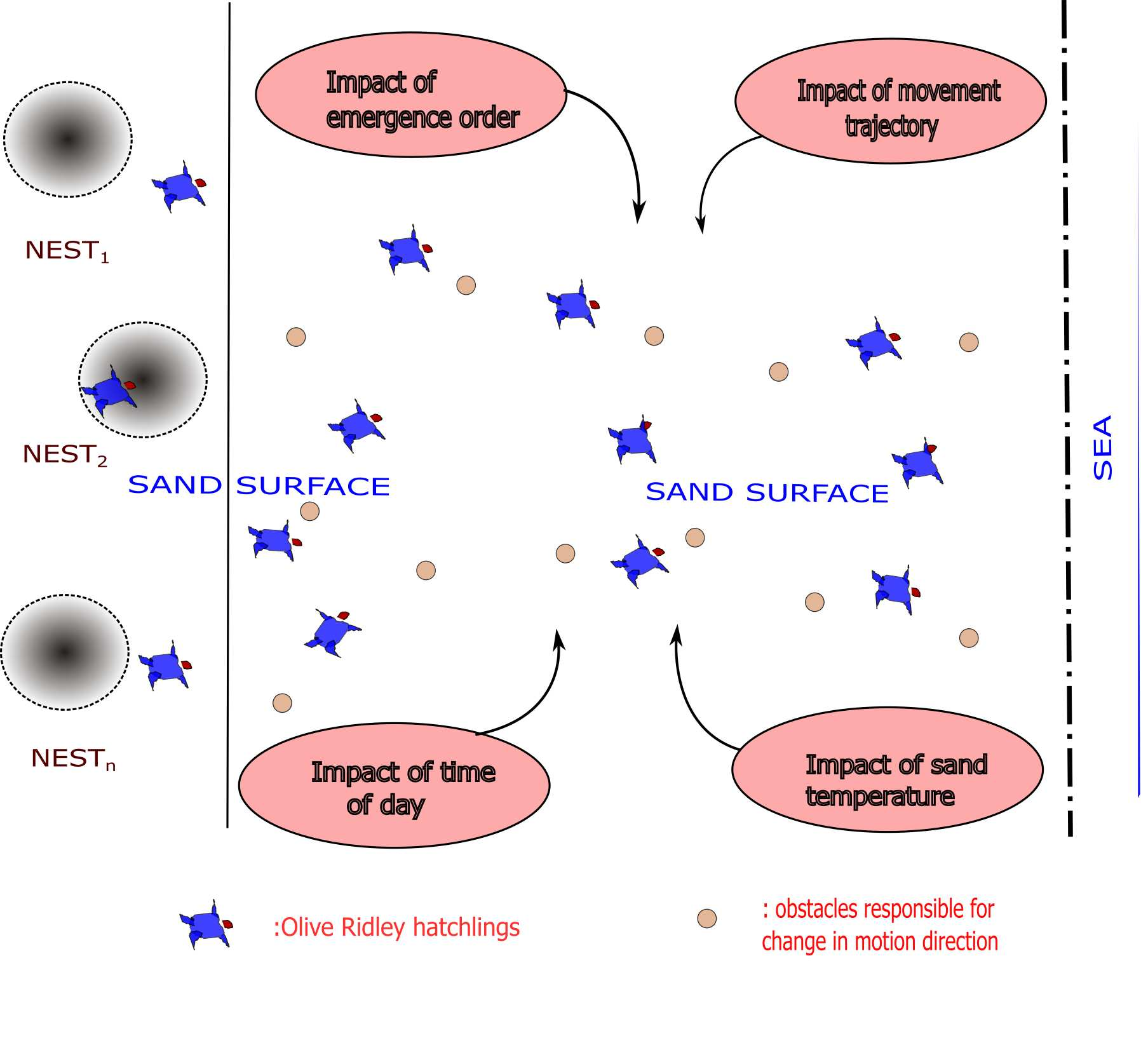}
    \caption{Schematic Diagram of proposed ORS}
    \label{fig_sch}
\end{figure}

\section{Mathematical Modelling}
This section presents in details the mathematical modelling of the proposed meta-heuristic scheme as as shown in Fig. \ref{fig_sch}. 
\subsection{Input Representation}
The input for ORS can be represented using a tuple of two dominant attributes, namely mass, the scalar quantity and velocity, the vector quantity. Thus, a hatchlings $h_{i}$ will be mathematically represented as given in Equation \ref{eq1}.
\begin{equation}
h_{i}= (m_{i}, \vec{v}_{i} )
\label{eq1}
\end{equation}
where $m_{i}=mass~ of~ i^{th} hatchling$, and $v_{i}=velocity~of~ i^{th}~ hatchling$. This representation logically helps in distinguishing a pool of hatchlings w.r.t their size and movement dynamics. In fact, this representation inherently signifies their fitness in terms of physical momentum to reach to the sea which can be mathematically expressed using Equation \ref{eq2}
\begin{equation}
f_{i}= m_{i} \cdot \vec{v}_{i} 
\label{eq2}
\end{equation}

The survival of hatchlings solely depends on how fast it can reach to the sea which is inherently dependent on its momentum expressed in Equation \ref{eq2}. Hence, any change in environmental factors and direction of movement will affect velocity, which in turn affects the fitness of the hatchlings.
\subsection{Initial Population Generation}
The initial population for ORS is generated using a population matrix where the mass and velocity vector will contain random values in the specified range. Formally, for a hatchlings of population size of $p$ and problem dimension $d$, it can be represented as a matrix shown in Equation \ref{eq3}.
\begin{equation}
H_{d} = 
\begin{pmatrix} 
(m_{1}, v_{11}) & \cdots & (m_{1}, v_{1d}) \\
\vdots & \ddots & \vdots \\ 
(m_{p}, v_{p1}) & \cdots & (m_{p}, v_{pd}) 
\end{pmatrix}
\label{eq3}
\end{equation}
\subsection{Environmental Impact}

According to background investigation presented in Section 2, the survival of a hatchlings depends on how fast it can reach to the sea. One important phenomenon which governs its momentum of movement to sea is environmental factors such as sand temperature, emergence order, and time of day \cite{r2}. This impact of individual factor can be expressed mathematically as follows. 

\subsubsection{Impact of sand temperature}
It is observed that as sand temperature, $S_{temp}$, increases, the hatchlings start crawling faster up to certain tolerable temperature, $temp_{tol}$, normally around $36^{\circ}C$-$37^{\circ}C$. After that, their momentum gets slower and when the temperature crosses a maximum temperature $temp_{max}$, above $40^{\circ}C$, they face thermal stress and most of the time, they get died. Hence, the velocity is set to $-\infty$ to mathematically model their death in case of temperature crossing maximum temperature.
\begin{equation}
\vec{v}^{t+1}= 
\begin{cases}
         \omega_{1} \vec{v}^{t},~~ if~~  S_{temp} \le temp_{tol}\\
         \frac{\vec{v}^{t}}{\omega_{2}}, ~~if~~  temp_{tol} < S_{temp} < temp_{max}\\
         -\infty,~~ if~~ S_{temp} > temp_{max}
 \end{cases}
 \label{eq4}
\end{equation}
Using Equation \ref{eq4}, the change in velocity at two time instances $t$ and $t+1$ due to sand temperature which affects the change in fitness is given by the Equation \ref{eq5}.
\begin{equation}
\Delta \vec{v}_{temp}= (\vec{v}^{t+1}-\vec{v}^{t})\\
=
\begin{cases}
         \vec{v}^{t}(\omega_{1}-1),~~ if~~ S_{temp} \le temp_{tol}  \\
         \frac {\vec{v}^{t}(1-\omega_{2})}{\omega_{2}}, ~~if~~  temp_{tol} < S_{temp} < temp_{max}\\
         -\infty,~~ if~~ S_{temp} > temp_{max}
 \end{cases}
 \label{eq5}
\end{equation}

\subsubsection{Impact of emergence order}
The emergence order of hatchlings plays a vital role in movement of the hatchlings towards sea. The hatchlings emerging early crawl faster than hatchlings emerging late. Hence, the velocity can be expressed as given in Equation \ref{eq6} according to their relative emergence order.  The visualization of emergence order is shown in Fig. 5.
\begin{figure} [th!]
    \centering
   \includegraphics[scale=0.6]{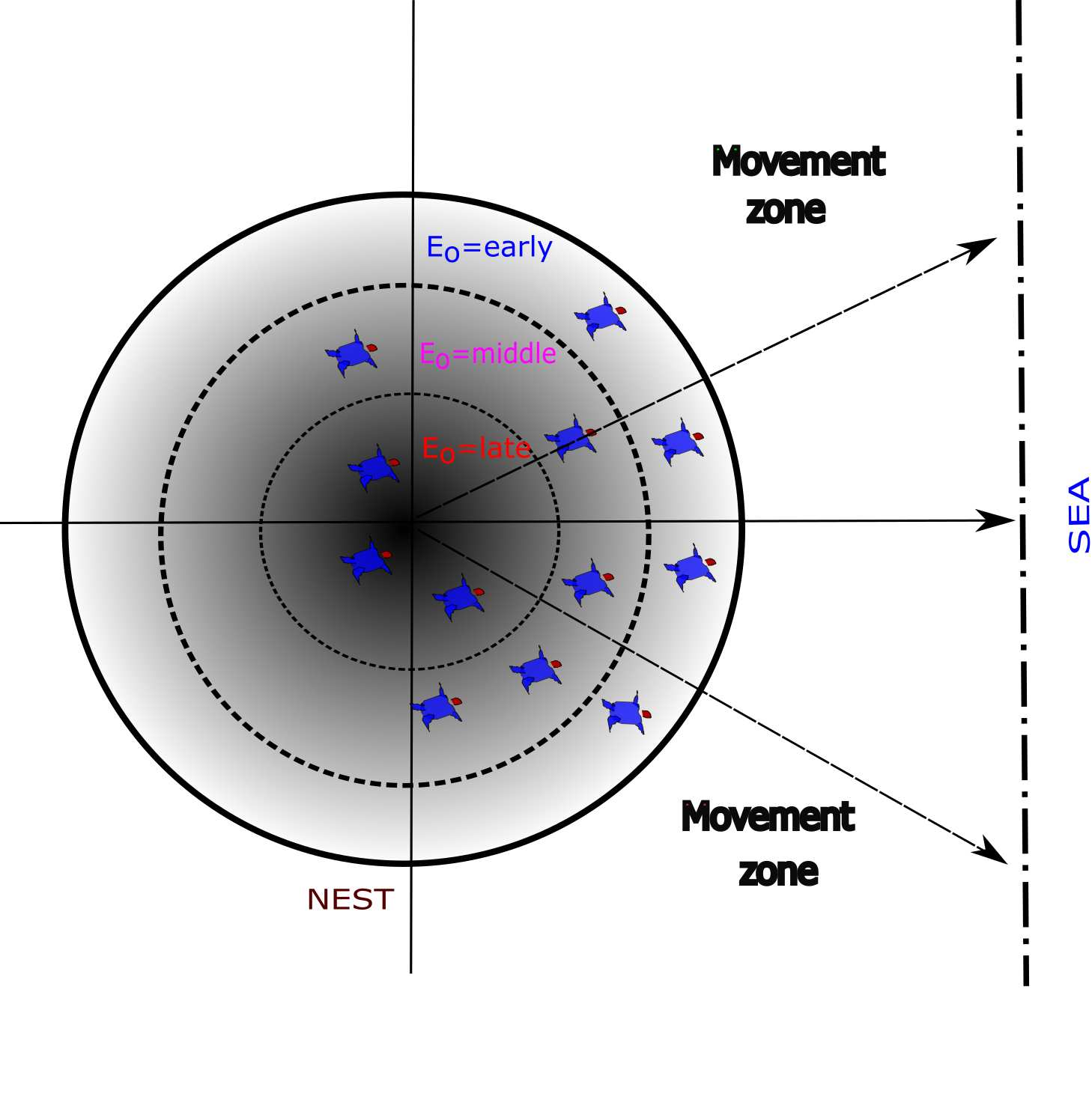}
    \caption{Emergence order of Olive Ridley hatchlings from nest}
\end{figure}
\begin{equation}
\vec{v}^{t+1}= 
\begin{cases}
         k \vec{v}^{t}+k_{1},~~ if~~  E_{o} = early\\
         k \vec{v}^{t},~~ if~~  E_{o} = middle\\
         k \vec{v}^{t}-k_{2},~~ if~~  E_{o} = late
\end{cases}
 \label{eq6}
\end{equation}
Using Equation \ref{eq6}, the change in velocity at different time instances due to emergence order which affects the change in fitness is given by the Equation \ref{eq7}.

\begin{equation}
\Delta \vec{v}_{emergence}= (\vec{v}^{t+1}-\vec{v}^{t})=
\begin{cases}
        \vec{v}^{t}(k-1)+k_{1},~~ if~~  E_{o} = early\\
        \vec{v}^{t}(k-1),~~ if~~  E_{o} = middle\\
        \vec{v}^{t}(k-1)-k_{2},~~ if~~  E_{o} = late
\end{cases}
 \label{eq7}
\end{equation}

\subsubsection{Impact of time of day}

Hatchlings mostly emerge from nest at night to protect themselves from heat. From dawn to around 8:00 AM ($t_{1}$ in Equation \ref{eq8}), they move faster as the day progresses. But, due to increase in sand temperature as the day progresses, their speed gets slower down from $t_{2}$ (12:00 noon). Again, from afternoon, $t_{3}$ till morning, they move faster. Since, the time of the day is indirectly affected by sand temperature, a similar model for impact due to time of the day is given in Equation \ref{eq8}.

\begin{equation}
\vec{v}^{t+1}= 
\begin{cases}
         \omega_{3} \vec{v}^{t},~~ if~~  t_{1} \le t_{d} < t_{2}\\
         \frac{\vec{v}^{t}}{\omega_{4}}, ~~if~~  t_{2} \le t_{d} < t_{3}\\
         \omega_{5} \vec{v}^{t},~~ if~~  t_{3}\le t_{d} < t_{1}
 \end{cases}
 \label{eq8}
\end{equation}
Using Equation \ref{eq8}, the change in velocity due to time of the day which affects the change in fitness is given by the Equation \ref{eq9}.
\begin{equation}
\Delta \vec{v}_{time}= (\vec{v}^{t+1}-\vec{v}^{t})\\
=
\begin{cases}
         \vec{v}^{t}(\omega_{3}-1),~~ if~~  t_{1} \le t_{d} < t_{2}  \\
         \frac {\vec{v}^{t}(1-\omega_{4})}{\omega_{4}}, ~~if~~  t_{2} \le t_{d} < t_{3}\\
         \vec{v}^{t}(\omega_{5}-1),~~ if~~  t_{3} \le t_{d} < t_{1}
 \end{cases}
 \label{eq9}
\end{equation}
Therefore, the overall environmental impact on momentum due to change in velocity can be modeled  by Equation \ref{eq10}. 
\begin{equation}
    \Delta \vec{v}_{env}= \Delta \vec{v}_{temp}+\Delta \vec{v}_{emergence}+\Delta \vec{v}_{time}
    \label{eq10}
\end{equation}
\begin{equation}
     r_{1}= p_{1}\Delta \vec{v}_{env}
    \label{eq11}
\end{equation}
In Equation \ref{eq11}, $p_{1}$ is a randomly assigned weight value to the overall environmental effect  in the range [0,1]. $r_{1}$ is a generated normalized value between 0 to 1. The $r_{1}$ value controls the fitness of the hatchlings due to overall impact of environmental factors.
\begin{figure}[th!]
  \centering
   \includegraphics[scale=0.8]{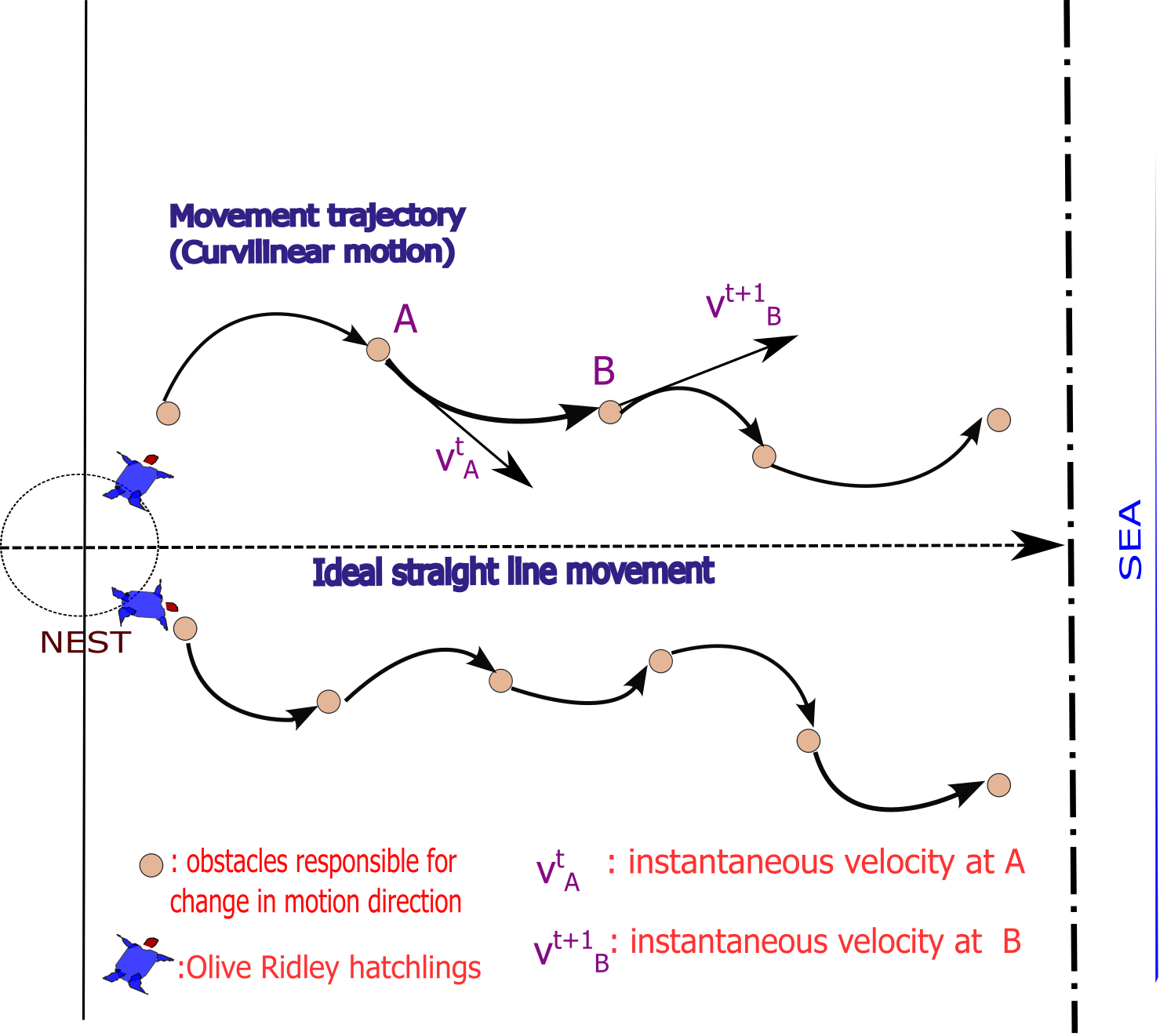}
   \caption{curvilinear trajectory movement of hatchling}
\end{figure}
\subsection{Movement Trajectory Impact}

Investigation on Olive Ridley hatchlings reveals that once the hatchlings are emerged from their nests, they start crawling towards sea. But due to different obstacles like debris on sea shore, they deviate from straight line movement. The deviation from straight line movement will insist a trajectory path which is modeled as a curvi-linear motion as shown in Fig. 6. As a consequence, it increases the probability of external attacks due to more time of exposure to the land predators and avian predators. This impact of trajectory movement on change in instantaneous velocity which affects the fitness of hatchlings can be modeled as follows.

\begin{figure}[h!]
  \centering
   \includegraphics[scale=0.8]{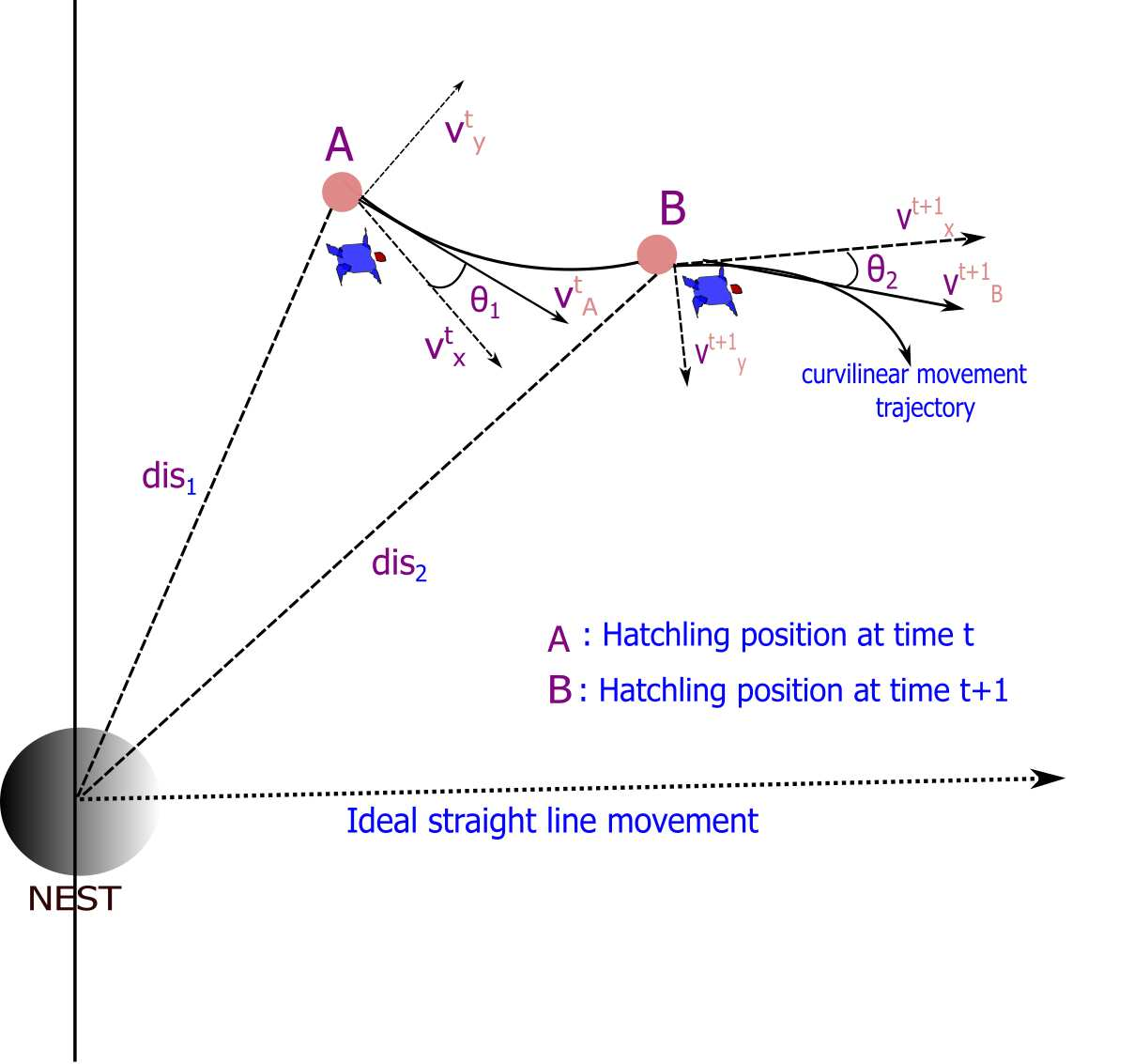}
   \caption{instantaneous velocity on trajectory of hatchling}
\end{figure}

In Fig. 6, the trajectory movement of a hatchling is realized using curvi-linear motion along the random obstacle  points.  The obstacle points are the  positions where the velocity of a hatchling changes. Considering, two reference points A and B on the trajectory as shown in Fig. 7, the respective instantaneous velocity $v_{A}^{t}$ and $v_{B}^{t+1}$ are given by Equation \ref{eq12} and \ref{eq13}. 

\begin{equation}
    v^{t}_{A}=\sqrt{(v_{x}^{t})^{2}+(v_{y}^{t})^{2}} ~~and~~\\
    \theta_{1} = \tan^{-1} (v_{y}^{t}/v_{x}^{t})
    \label{eq12}
\end{equation}

\begin{equation}
    v^{t+1}_{B}=\sqrt{(v_{x}^{t+1})^{2}+(v_{y}^{t+1})^{2}} ~~and~~\\
    \theta_{2} = \tan^{-1} (v_{y}^{t+1}/v_{x}^{t+1})
    \label{eq13}
\end{equation}

From Equation \ref{eq12} and \ref{eq13}, the change in tangential velocity due to movement trajectory is given by Equation \ref{eq14}.
\begin{equation}
\begin{split}
\Delta \vec{v}_{trajectory}= (\vec{v}_{B}^{t+1}-\vec{v}_{A}^{t})\\
=\sqrt{(v_{x}^{t+1})^{2}+(v_{y}^{t+1})^{2}}-\sqrt{(v_{x}^{t})^{2}+(v_{y}^{t})^{2}}
\end{split}
  \label{eq14}
\end{equation}
Further, the angular change in velocity due to trajectory movement is given by Equation \ref{eq15}.
\begin{equation}
\begin{split}
\Delta \theta_{trajectory}= \theta_{2}-\theta_{1} \\
= \tan^{-1} (v_{y}^{t+1}/v_{x}^{t+1})-\tan^{-1} (v_{y}^{t}/v_{x}^{t})
\end{split}
  \label{eq15}
\end{equation}

Hence, the controlling factor on fitness of a hatchling due to movement trajectory is computed using Equation \ref{eq16}.
\begin{equation}
    r_{2}= p_{2}\Delta \vec{v}_{trajectory}
    \label{eq16}
\end{equation}

In Equation \ref{eq16}, $p_{2}$ is randomly assigned value in the range [0,1].$r_{2}$ controls the fitness of the hatchlings due to impact of movement trajectory factor.

\subsection{Fitness Update Policy}

The resultant impact on change of fitness due to both environment and movement trajectory can be computed using Equation \ref{17} as follows:

\begin{equation}
\Delta \vec{v}_{resultant}= (r_{1}+r_{2}) 
\label{17}
 \end{equation}

Further, a parameter survival factor for each hatchling is defined as follows.
\textbf{Survival Factor}: The survival factor $S_{f}$ of a hatchling $i$ in $t^{th}$ generation is computed using Equation \ref{eq18}.
\begin{equation}
    S^{t}_{f}(i)= \frac{f^{t}_{max}-f^{t}(i)}{f^{t}_{max}-f^{t}_{min}}
    \label{eq18}
\end{equation}
Hence, the fitness update at $(t+1)^{th}$ iteration  will be done as per the Equation \ref{eq19}. 

\begin{equation}
f_{i}^{t+1} =
\begin{cases}
m_{i}\{(\vec{v}_{i}^{t}+ \Delta \vec{v}_{resultant})+ \vec{v}^{t}_{best}\}, ~if~ S^{t}_{f}(i)< 0.3\\
m_{i}\{(\vec{v}_{i}^{t}- \Delta \vec{v}_{resultant})+ \vec{v}^{t}_{best}\}, otherwise
\end{cases}
\label{eq19}
\end{equation}

\section{The ORS Optimizer}

This section presents the description of ORS strategy and pseudo code of ORS in Algorithm 1 using above discussed mathematical modelling. The algorithm can be summarized as below. 
\subsection{Algorithm Description}
\SetKwComment{Comment}{/* }{ */}
\RestyleAlgo{ruled}
\begin{algorithm}
\caption{ORS Optimizer}\label{alg:two}
\KwData{$Hatchling~ population~ H= \{h_{1},h_{2},...,h_{n}\}, iter_{max}$}
\KwResult{$Optimal~ hatchling, h_{opt}$}
$i \gets 1$\;
$iter \gets 1$\;
\While{$i \le  n$}
{
$Initialize~ m_{i}~and~ v^{0}_{i}$\;
$f_{i} \gets m_{i}* v^{0}_{i}$ \Comment*[r] {fitness function using Equation \ref{eq2}}
$i \gets i+1$\;
}
\While{$i \le  n$}
{
$Compute~survival~factor~S^{0}_{f}(i)$\;
$i \gets i+1$\;
}
$Store~ hatchling~ h^{opt} ~with~ optimal f_{i}$\;
\While{$iter \le iter_{max}$}
{
\While{$i \le n$}
{
$Compute~ \Delta \vec{v}^{i}_{env}~and~ \Delta \vec{v}^{i}_{trajectory}$ \Comment*[r] {using Equation \ref{eq10} and \ref{eq14}}
$p_{1} \gets random(0,1)$\;
$p_{2} \gets random(0,1)$\;
$r_{1} \gets p_{1}\Delta \vec{v}^{i}_{env}$\;
$r_{2} \gets p_{2}\Delta \vec{v}^{i}_{trajectory}$\;
$\Delta \vec{v}^{i}_{resultant} \gets (r_{1}+r_{2}) $\;
\eIf{$S^{iter-1}_{f}(i) \le 0.3$}
{
$f_{i}^{iter}=m_{i}\{(\vec{v}_{i}^{iter-1}+ \Delta \vec{v}_{resultant})+ \vec{v}^{iter-1}_{best}\}$\;
}
{
$f_{i}^{iter}=m_{i}\{(\vec{v}_{i}^{iter-1}- \Delta \vec{v}_{resultant})+ \vec{v}^{iter-1}_{best}\}$\;
}
$Compute~and~store~survival~factor~S^{iter}_{f}(i)$\;
}
\If{$f^{iter}_{i} < f(h_{opt})$}
{
$h_{opt}=h_{i}$\;
}
$iter \gets iter+1$\;
}
$Display~optimal~hatchling~~h_{opt}$\;
\end{algorithm}
The algorithm takes input as population of hatchlings $H$ which are represented as tuples as shown in Equation \ref{eq1}. The output will be the optimal hatchling $h_{opt}$ which optimizes the objective function of an underlying problem. The algorithm starts with initialization of mass and initial velocities $v^{0}$ of each hatchling $i$. The initial survival factor of each hatchling is computed using Equation \ref{eq18} and the best hatchling with optimal fitness value is stored.The algorithm is iterated for a maximum number of iterations $iter_{max}$ and for every hatchling the respective impact of environmental factors $\Delta \vec{v}_{env}$ and movement trajectory $\Delta \vec{v}_{trajectory}$ is computed using Equation \ref{eq11} and \ref{eq14}. The resultant impact $\Delta \vec{v}_{resultant}$ is then calculated. For hatchlings which have lesser survival factors than a threshold value (value considered is 0.3), the fitness is updated by adding resultant impact with previous velocity value and tuned towards best velocity value. Otherwise, it is subtracted and tuned. 

\subsection{Algorithm Analysis}
This section presents theoretical analysis of ORS algorithm like complexity analysis, exploitation analysis, exploration analysis, and convergence analysis. 
\subsubsection{Complexity Analysis}
In this section, the computational complexity of the proposed ORS optimizer (Algorithm 1) is computed by counting the number of basic operations. The algorithm is supplied with the inputs $n$(No. of hatchlings) and $iter_{max}$ (maximum iteration). The setting of the initial population takes $O(n)$ time. The subsequent step computes the survivability factor of the initial population with a $O(n)$ time requirement. After this, the hatchlings undergo successive iterations of modifications that require $O(iter_{max}*n)$ time. Therefore, the computational complexity of this algorithm is   $O(n) + O(n) + O(iter_{max}*n)=O(iter_{max}*n)$.
  
\subsubsection {Exploitation Analysis}
The movement trajectory of the hatchlings determines the exploitation when a hatchling tries to reach the sea as soon as possible. As the movement trajectory is modeled by two-dimensional curvilinear motion, the magnitude and direction of the velocity vector play a major role in defining the optimality of the solution. These values are modified respectively as per the fittest hatchlings to attain optimality. Also, each hatchling tries to increase its fitness to enhance the chance of survivality.

\subsubsection{Exploration Analysis}
The three environmental factors: send temperature, emergence order, and time of the day cause hatchlings to deviate from their original path but still enhance the change of survivability by exploring different paths. It directs some of the non-fit hatchlings that are deviating from the optimal path to become the fittest hatchings.

\subsubsection{Convergence Analysis}
The convergence analysis describes the consensus between the exploration and exploitation to achieve global optimum. In the proposed ORS optimizer, the curvilinear motion of the hatchlings moving out of the nest (representing exploitation) and going toward the sea is regulated by three environmental factors: sand temperature, emergence order, and time of day (representing exploration). The survival factor determines the fitness of the hatchlings and decides the next population. This process of elimination enables the selection of the best solutions/hatchlings to the next population hence ensuring the convergence after several iterations at an optimal or near-optimal trade-off point of exploration and exploitation.

\section{Simulation Results \& Discussion}

The simulation is performed using MATLAB tool on a machine of 8 Gb RAM, core-i3 3.0 GHz Processor Speed, and Windows-10 64-bit Operating System. The number of simulation run taken is 100 with 1000 iterations. The performance of ORS algorithm is analyzed by comparing it with standard meta-heuristics approaches such as TSA \cite{19}, MVO \cite{9}, SCA \cite{10}, GWO \cite{4}, WOA \cite{11}, BA \cite{2}, and DE \cite{1}. The performance is compared using two important metrics those are Mean (M) and Standard Deviation (SD). All these algorithms are run and tested over 14 standard Benchmark Functions, these are also applied and used by many of the state-of-the-art meta-heuristic \cite{19,9,10,4,11,2,1} for the performance analysis. From 14 Benchmark functions, 8 functions ($Fn_{1}$, $Fn_{3}$, $Fn_{5}$, $Fn_{8}$, $Fn_{9}$, and $Fn_{10}$) are from standard test suite of  CEC 2005 \cite{suganthan2005problem}, CEC 2008 \cite{tang2008benchmark}, and CEC 2010 \cite{tang2010benchmark}.\par
For better analysis we have considered unimodal functions ($Fn_1$-$Fn_7$), multimodal functions ($Fn_8$-$Fn_{12}$), and fixed dimension multimodal functions ($Fn_{13}$-$Fn_{14}$). The complete representation of these functions are shown in Table 2 and Figures 8-21. This table also have three more columns representing Dimension, Range and $F_{min}$. Dimension signify the functions dimension, Range denotes the space of search in functions boundary, and $F_{min}$ shows the optimum value. Also, to check the efficacy of ORS on recent complex benchmark functions,  test suite of CEC-C06 2019 \cite{abdullah2019fitness} is used for performance comparison of ORS with other state-of-the-art metaheuristic. This test suite contains 10 complex functions where the optimum value for all these functions is 1. The results of this is shown in Table 7 and Fig. 22-31, where it is found that ORS performs better than other metaheuristic algorithms in CEC01 and CEC02. In CEC03 and CEC10, ORS shows at par results with other metaheuristic. However, from CEC04-CEC09, the behavior of ORS algorithm is sub-optimal when compared with other metaheuristics. 
\begin{landscape}
\begin{table}[h]
\caption{Benchmark Functions taken for performance analysis}
\scriptsize
\begin{tabular}{lllll}
\hline
Sl. No. & Benchmark Function                                                                                                                & Dimension & Range Taken      & $Fn_{min}$ \\ \hline
1 & $Fn_{p} = \sum_{j=1}^{n} p_j^2$                                                                                                    & 30        & \{ -100, 100 \}  & 0          \\
2 & $Fn_{p} = \sum_{j=1}^{n} \left| p_j \right| + \prod_{j=1}^{n} \left| p_j \right|$                                                                                                       & 30        & \{ -10, 10 \}    & 0          \\
3 & $Fn_{p} = \sum_{j=1}^{n} \left( \sum_{k=1}^{j} p_k \right)^2$                                                                                                                                & 30        & \{ -100, 100 \}  & 0          \\
4 & $Fn_{p} = \max_{1 \leq j \leq p} \left\{ p_{j} \right\}$                                                                                                                        & 30        & \{ -100, 100 \}  & 0          \\
5 & $Fn_{p} = \sum_{j=1}^{p-1} \left[ 100 \left ( p_{j+1} - p_j^2 \right )^2 + \left ( p_j - 1 \right )^2 \right ]$                                                                                                                                     & 30        & \{ -30, 30 \}    & 0          \\
6 & $Fn_{p} = \sum_{j=1}^{p} \left[ p_j + 0.5 \right]^2$                                                                                                                                                                             & 30        & \{ -100, 100 \}   & 0          \\
7 & $Fn_{p} = \sum_{j=1}^{p} j p_j^4 + \text{random}\left[0,1\right]$                                                                                                                                 & 30        & \{ -1.28, 1.28 \} & 0          \\
8 & $Fn_{p} = \sum_{j=1}^{p} \left[ p_j^2 - 10 \cos \left( 2 \pi p_j \right) + 10 \right]$                                                                                                        & 30        & \{ -5.12, 5.12 \} & 0          \\
9 & $Fn_{p} = -20 \exp \left( -0.2 \sqrt{\frac{1}{n} \sum_{j=1}^{n} p_j^2} \right) - \exp \left( \frac{1}{n} \sum_{j=1}^{n} \cos \left( 2 \pi p_j \right) \right) + 20 + e$                                                                                                                                           & 30        & \{ -32, 32 \}     & 0          \\
10 & $Fn_{p} = \frac{1}{4000} \sum_{j=1}^{n} p_j^2 - \prod_{j=1}^{n} \cos \left( \frac{p_j}{\sqrt{j}} \right) + 1$                                                                                                                                          & 30        & \{ -600, 600 \}   & 0          \\
11 & \begin{tabular}[c]{@{}l@{}}$Fn_{p} = \frac{\pi}{n} \left\{ 10 \sin \left( \pi q_1 \right) + \sum_{j=1}^{n-1} \left( q_j - 1 \right)^2 \left[ 1 + 10 \sin^2 \left( \pi q_{j+1} \right) \right] + \left( q_n - 1 \right)^2 \right\} +$ \\ 
$\sum_{j=1}^{n} u \left( p_j, 10, 100, 4 \right)$ \\ 
$q_j = 1 + \frac{p_j + 1}{4}$ \\ 
$u \left( p_j, a, k, m \right) = \left\{ \begin{matrix}
k (p_j - a)^m & \text{if } p_j > a \\ 
0 & \text{if } -a < p_j < a \\ 
k (-p_j - a)^m & \text{if } p_j < -a
\end{matrix} \right.$\end{tabular} & 30        & \{ -50, 50 \}     & 0          \\
12 & $Fn_{p} = 0.1 \left\{ \sin^2 (3 \pi p_1) + \sum_{j=1}^{n} (p_j - 1)^2 \left[ 1 + \sin^2 (3 \pi p_j + 1) \right] + (p_n - 1)^2 \left[ 1 + \sin^2 (2 \pi p_j) \right] \right\} + \sum_{j=1}^{n} u \left( p_j, 5, 100, 4 \right)$                                                                                                            & 30        & \{ -50, 50 \}     & 0          \\
13 & $Fn_{p} = \sum_{j=1}^{11} \left[ a_i - \frac{p_1 (b_j^2 + b_j p_2)}{b_j^2 + b_j p_3 + p_4} \right]$                                                                                                                                                    & 4         & \{ -5, 5 \}       & 0.00030    \\
14 & $Fn_{p} = \left(p_2 - \frac{5.1}{4 \pi^2} p_1^2 + \frac{5}{\pi} p_1 - 6\right)^2 + 10 \left(1 - \frac{1}{8 \pi}\right) \cos p_1 + 10$                                                                                                 & 2         & \{ -5, 5 \}       & 0.398  
\\ \hline
\end{tabular}
\end{table}
\end{landscape}
\subsection{Exploitation Analysis}
From the results shown in Table 3 and Table 4, that show the Mean and Standard Deviation, it is found that ORS outperforms TSA, MVO, SCA, GWO, WOA, BA, and DE. The mean and standard deviation of ORS is lesser than the above algorithms. The results are more optimum and exploited. The main reason for this best exploitation is use of novel movement trajectory operator. 

\subsection{Exploration Analysis}

The multimodal functions used have many local optima and the number increases exponentially when the dimension increases. From the results of multimodal functions ($Fn_8$-$Fn_{12}$) shown in Table 3 and Table 4, it is found that ORS outperforms TSA, MVO, SCA, GWO, WOA, BA, and DE in terms of Mean and Standard Deviation. So, ORS achieves exploration as per the above result. 

\begin{table}[h!]
\caption{Mean result for Benchmark functions}
\scriptsize
\begin{tabular}{lllllllll}
\hline
          & ORS      & TSA      & MVO      & SCA      & GWO      & WOA      & BA       & DE       \\ \hline
$Fn_1$    & 0.121775 & 341.1036 & 2553.003 & 10852.86 & 231.0256 & 231.8546 & 27716.02 & 1594.328 \\ 
$Fn_2$    & 0.006654 & 3.41E+08 & 1.98E+08 & 4518491  & 1.51E+08 & 1.42E+08 & 19349671 & 79322008 \\ 
$Fn_3$    & 0.643968 & 766.6429 & 6890.685 & 28937.39 & 944.163  & 30275.12 & 64032.12 & 32853.08 \\ 
$Fn_4$    & 0.004699 & 2.908256 & 16.77825 & 45.77156 & 1.089081 & 25.57383 & 62.17692 & 21.80284 \\ 
$Fn_5$    & 37.89641 & 649027.1 & 2787584  & 62374054 & 555094.4 & 633031.2 & 46261854 & 3508269  \\ 
$Fn_6$    & 5.167435 & 381.7262 & 2554.495 & 10260.19 & 232.3392 & 210.7077 & 23635.84 & 1571.587 \\ 
$Fn_7$    & 0.010584 & 0.344007 & 1.368568 & 22.73343 & 0.253875 & 0.374043 & 46.13789 & 1.665227 \\ 
$Fn_8$    & 0.154612 & 165.5043 & 217.0572 & 117.2501 & 11.72901 & 10.17045 & 357.5302 & 104.5092 \\ 
$Fn_9$    & 0.009408 & 3.117101 & 7.824604 & 17.56314 & 0.271692 & 0.21832  & 19.78864 & 2.406784 \\ 
$Fn_{10}$ & 0.006181 & 3.494719 & 24.11361 & 98.3579  & 2.199661 & 2.043611 & 260.4895 & 14.76768 \\ 
$Fn_{11}$ & 0.797094 & 1208216  & 3690831  & 1.51E+08 & 1082768  & 1181871  & 1.04E+08 & 5861192  \\ 
$Fn_{12}$ & 2.977152 & 2650070  & 9937908  & 3.05E+08 & 2543933  & 2381996  & 2.43E+08 & 13889939 \\ 
$Fn_{13}$ & 0.00093  & 0.002888 & 0.001229 & 0.001347 & 0.006492 & 0.000891 & 0.015872 & 0.000966 \\ 
$Fn_{14}$ & 0.505752 & 0.398987 & 0.40105  & 0.405015 & 0.399398 & 0.398714 & 0.486164 & 0.399864 \\ \hline
\end{tabular}
\end{table}

\begin{table}[h]
\caption{Standard Deviation result for Benchmark functions}
\scriptsize
\begin{tabular}{lllllllll}
\hline
          & ORS      & TSA      & MVO      & SCA      & GWO      & WOA      & BA       & DE       \\ \hline
$Fn_1$    & 0.138611 & 52.03758 & 93.46913 & 3567.339 & 19.26235 & 34.16675 & 10304.33 & 78.16635 \\ 
$Fn_2$    & 0.005074 & 4.92E+08 & 3.67E+08 & 14242246 & 3.99E+08 & 2.85E+08 & 35725932 & 1.28E+08 \\ 
$Fn_3$    & 0.916523 & 196.9474 & 922.9516 & 3192.09  & 225.0597 & 5050.377 & 16460.16 & 2563.732 \\ 
$Fn_4$    & 0.00374  & 0.851034 & 1.327478 & 4.983255 & 0.138747 & 19.71507 & 4.949388 & 1.270501 \\ 
$Fn_5$    & 8.440492 & 117634.2 & 309040   & 10096225 & 113212.2 & 152410.5 & 16717180 & 528584.9 \\ 
$Fn_6$    & 1.796612 & 89.96231 & 112.7935 & 2315.28  & 23.93805 & 30.09364 & 5124.935 & 42.77381 \\ 
$Fn_7$    & 0.000259 & 0.114226 & 0.153236 & 4.401741 & 0.020152 & 0.073871 & 27.54735 & 0.166331 \\ 
$Fn_8$    & 0.103571 & 31.7317  & 28.39093 & 23.63758 & 2.455645 & 4.182867 & 24.06147 & 1.908624 \\ 
$Fn_9$    & 0.005309 & 1.370914 & 0.128569 & 4.090547 & 0.015907 & 0.023548 & 0.307332 & 0.053016 \\ 
$Fn_{10}$ & 0.00313  & 0.396741 & 1.179602 & 24.05511 & 0.379835 & 0.252245 & 70.07441 & 0.795249 \\ 
$Fn_{11}$ & 0.115345 & 406707.1 & 860502.3 & 27700489 & 242808.7 & 535531.9 & 83305784 & 1111808  \\ 
$Fn_{12}$ & 0.052532 & 719337.7 & 2371328  & 50149521 & 535178.9 & 434561.9 & 1.3E+08  & 2342402  \\ 
$Fn_{13}$ & 0.000416 & 0.006205 & 0.000447 & 0.000315 & 0.009643 & 0.000561 & 0.016428 & 0.000127 \\ 
$Fn_{14}$ & 0.105807 & 0.001084 & 0.00236  & 0.00359  & 0.000646 & 0.00091  & 0.203315 & 0.00124  \\ \hline
\end{tabular}
\end{table}

\begin{figure}[h]
    \centering
    \includegraphics[width=16cm]{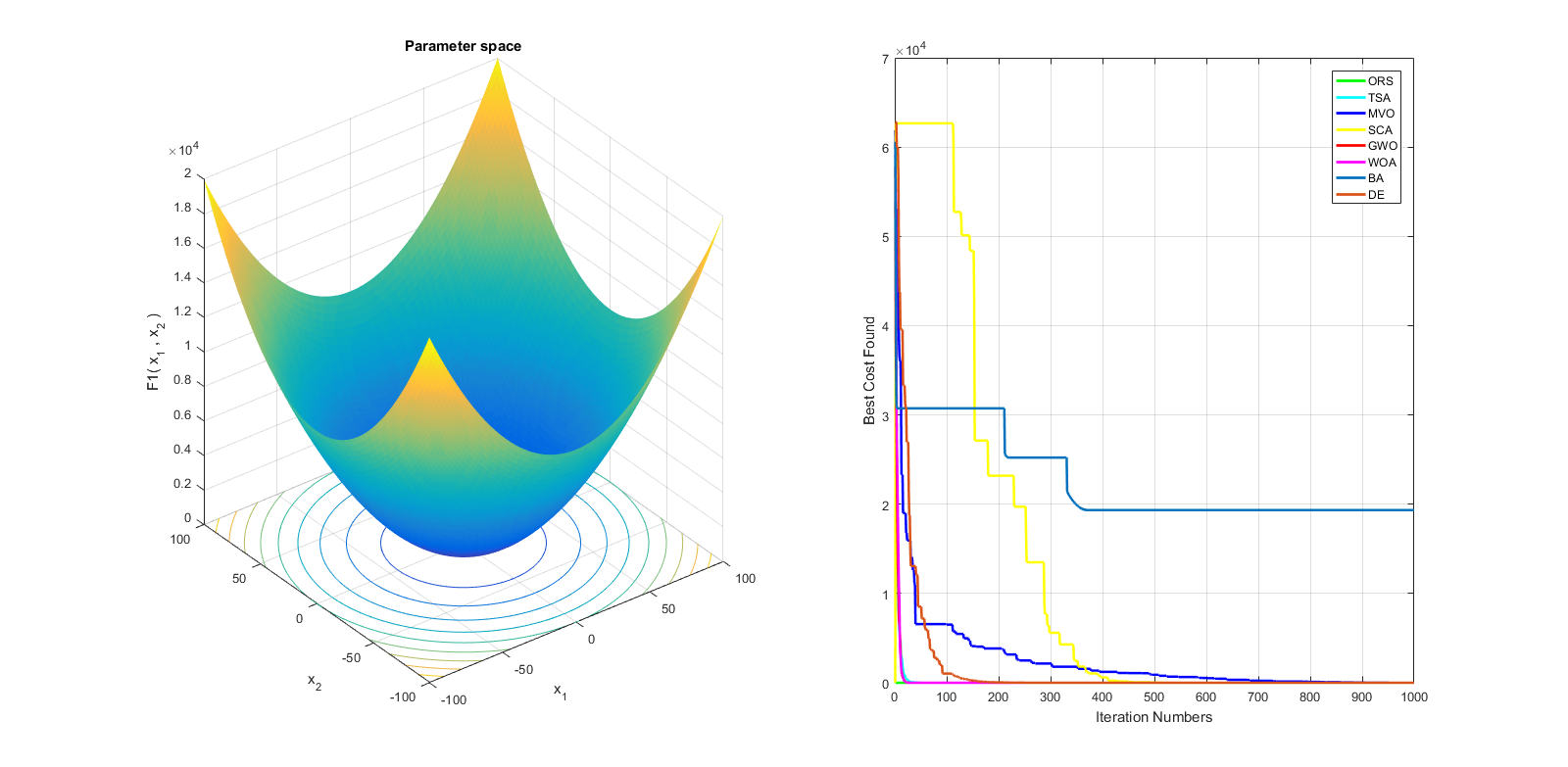}
    \caption{Representation of $Fn_1$ Benchmark function.}
    \label{fig:galaxy}
\end{figure}
\begin{figure}[h]
    \centering
    \includegraphics[width=16cm]{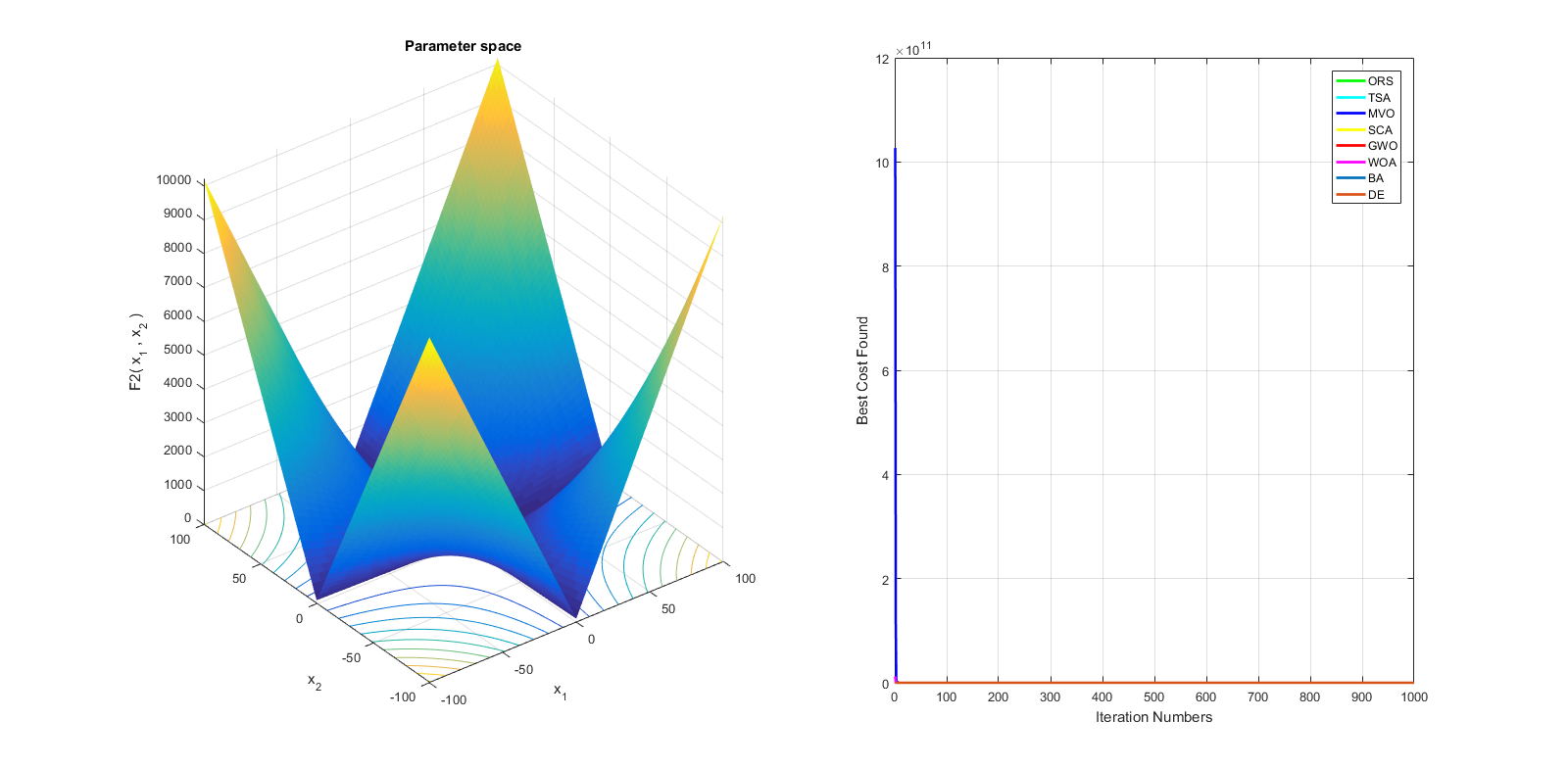}
    \caption{Representation of $Fn_2$ Benchmark function.}
    \label{fig:galaxy}
\end{figure}
\begin{figure}[h]
    \centering
    \includegraphics[width=16cm]{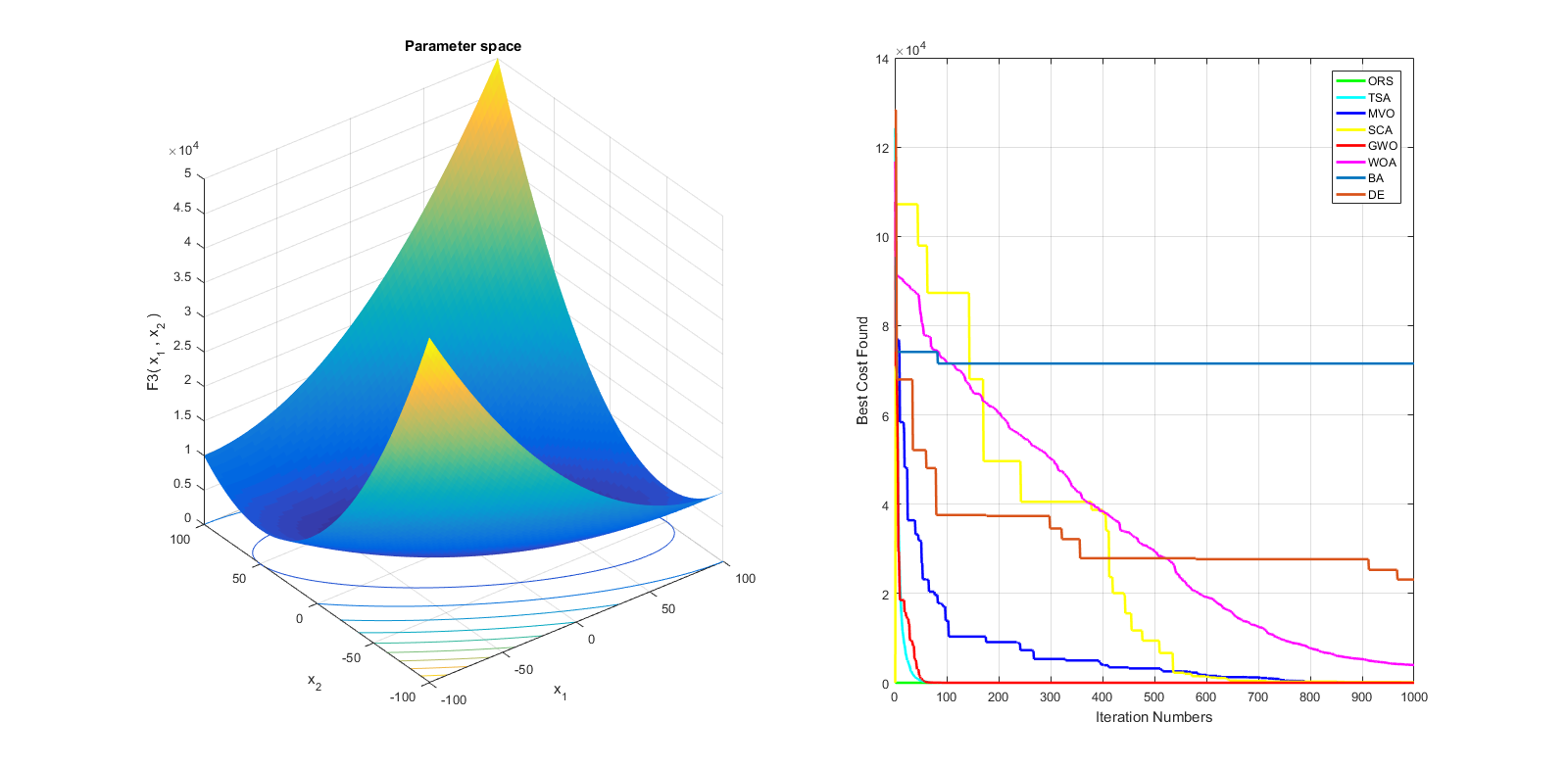}
    \caption{Representation of $Fn_3$ Benchmark function.}
    \label{fig:galaxy}
\end{figure}
\begin{figure}[h]
    \centering
    \includegraphics[width=16cm]{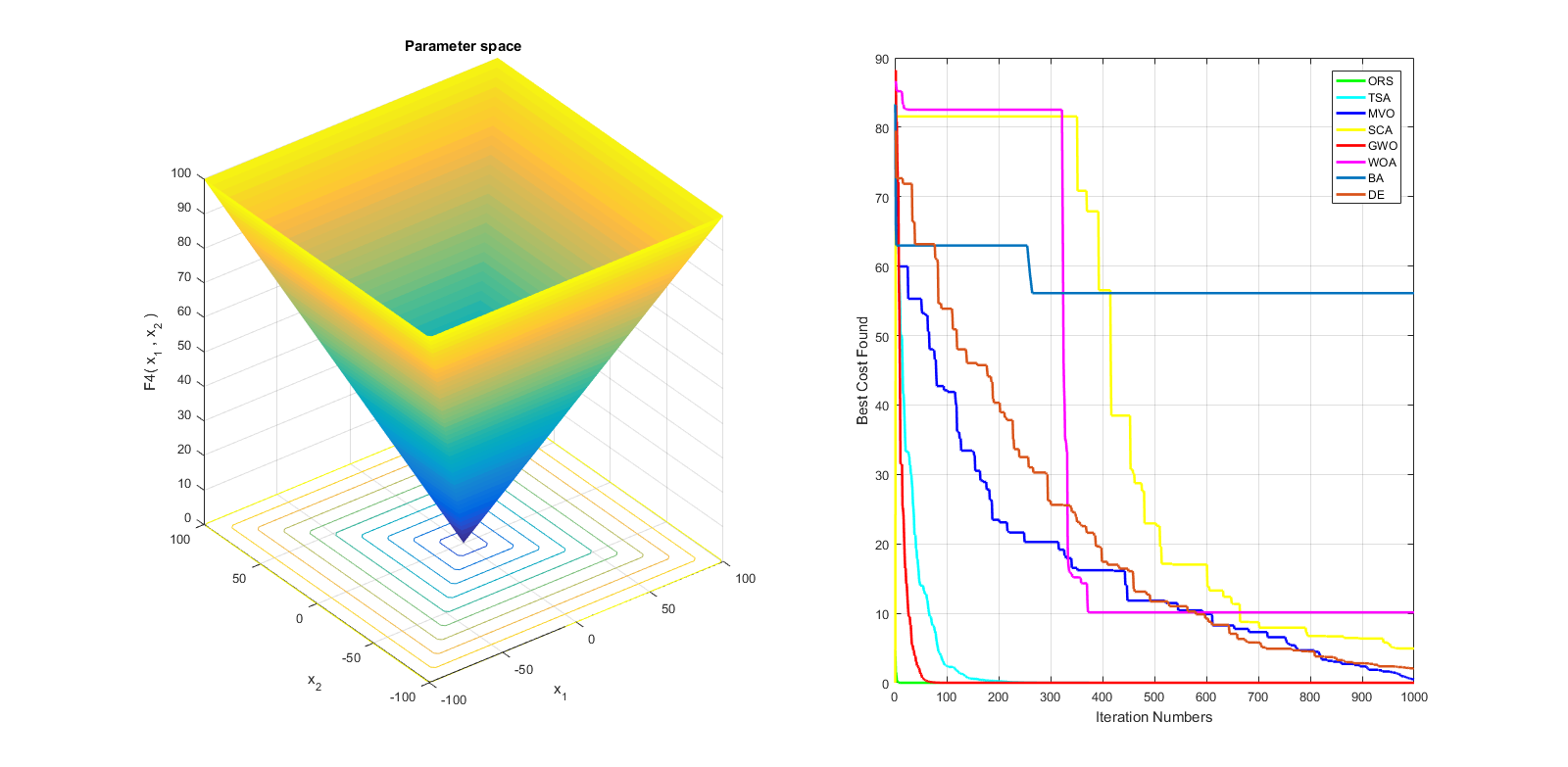}
    \caption{Representation of $Fn_4$ Benchmark function.}
    \label{fig:galaxy}
\end{figure}
\begin{figure}[h]
    \centering
    \includegraphics[width=16cm]{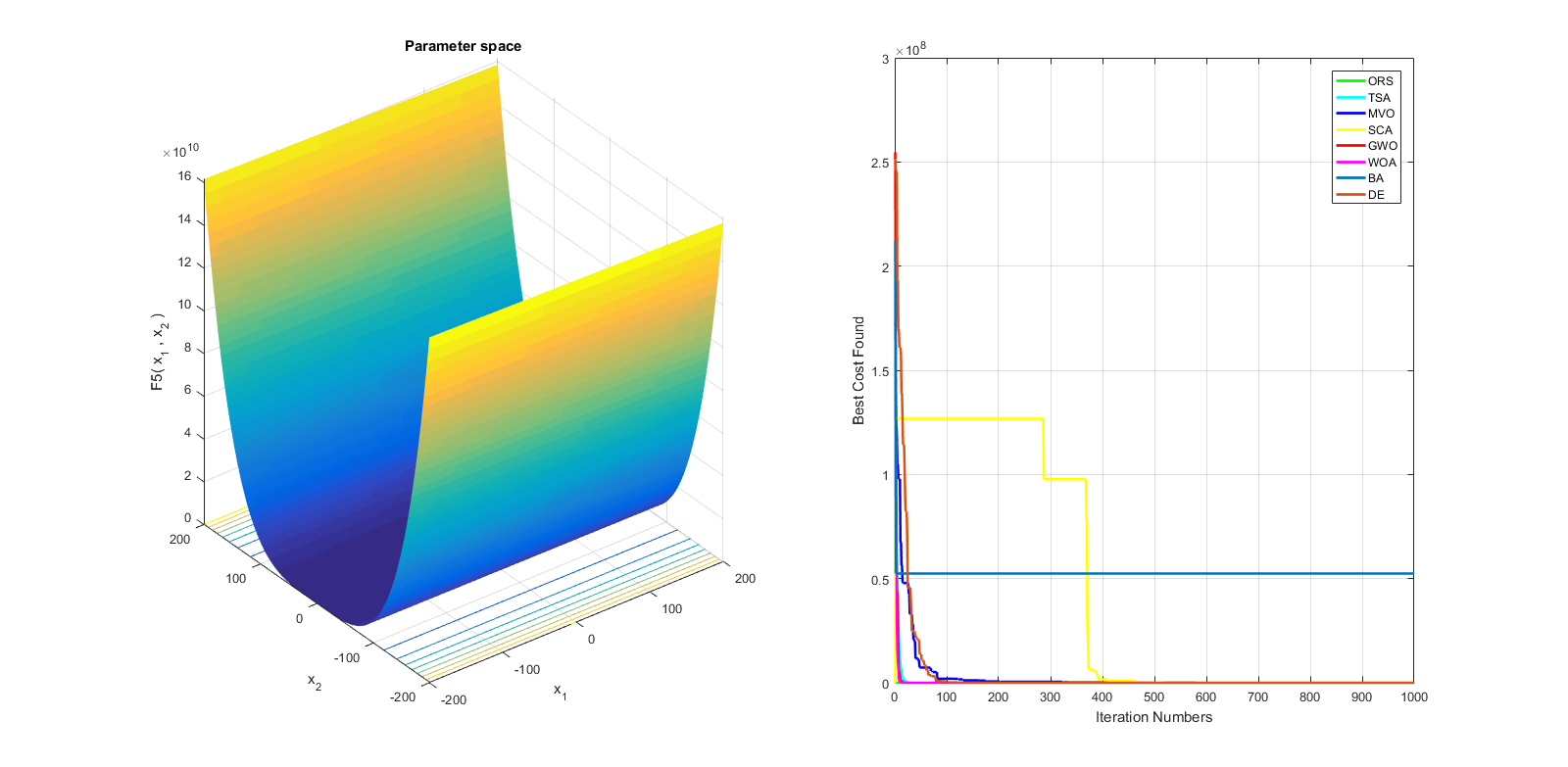}
    \caption{Representation of $Fn_5$ Benchmark function.}
    \label{fig:galaxy}
\end{figure}
\begin{figure}[h]
    \centering
    \includegraphics[width=16cm]{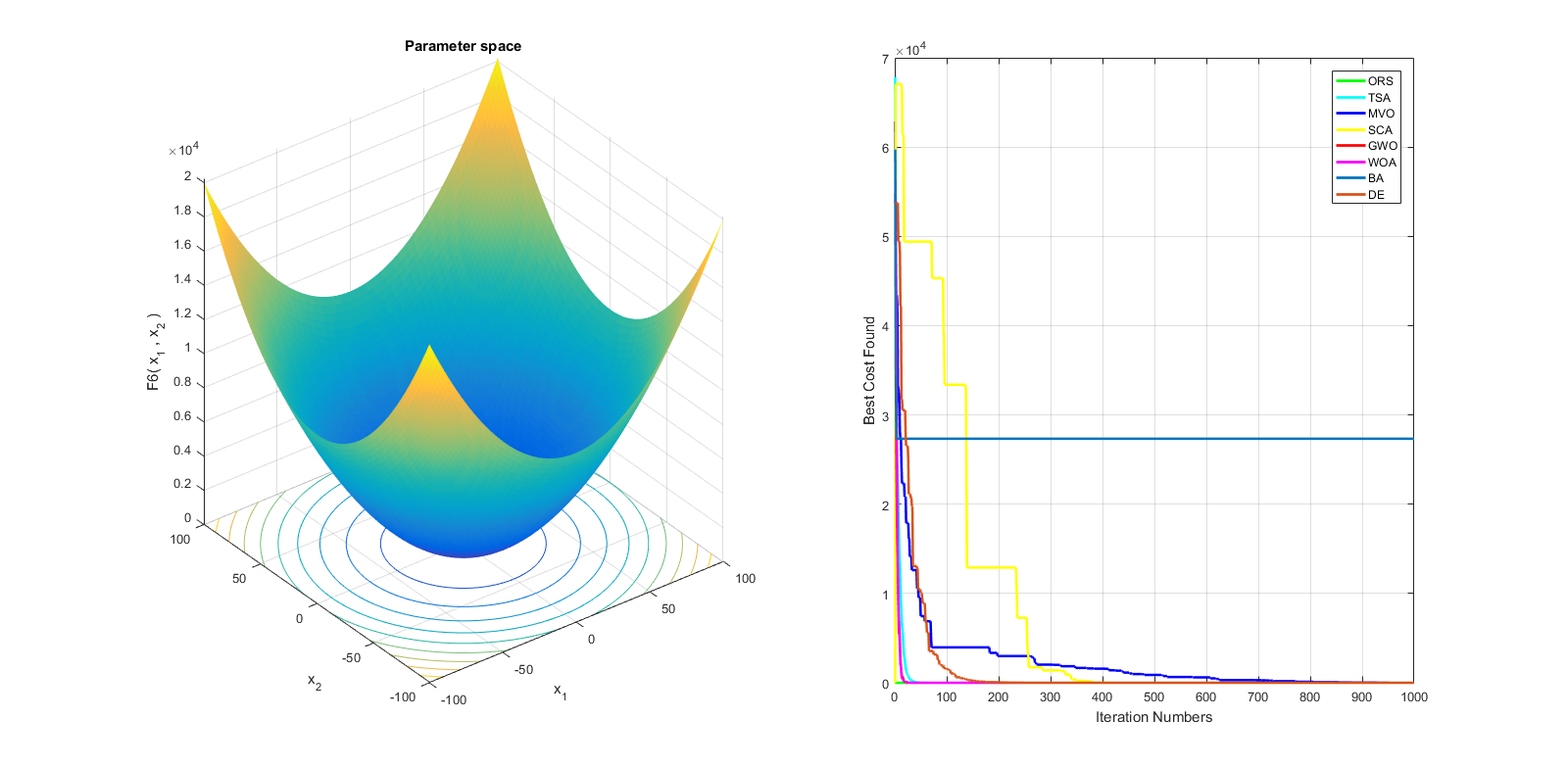}
    \caption{Representation of $Fn_6$ Benchmark function.}
    \label{fig:galaxy}
\end{figure}
\begin{figure}[h]
    \centering
    \includegraphics[width=16cm]{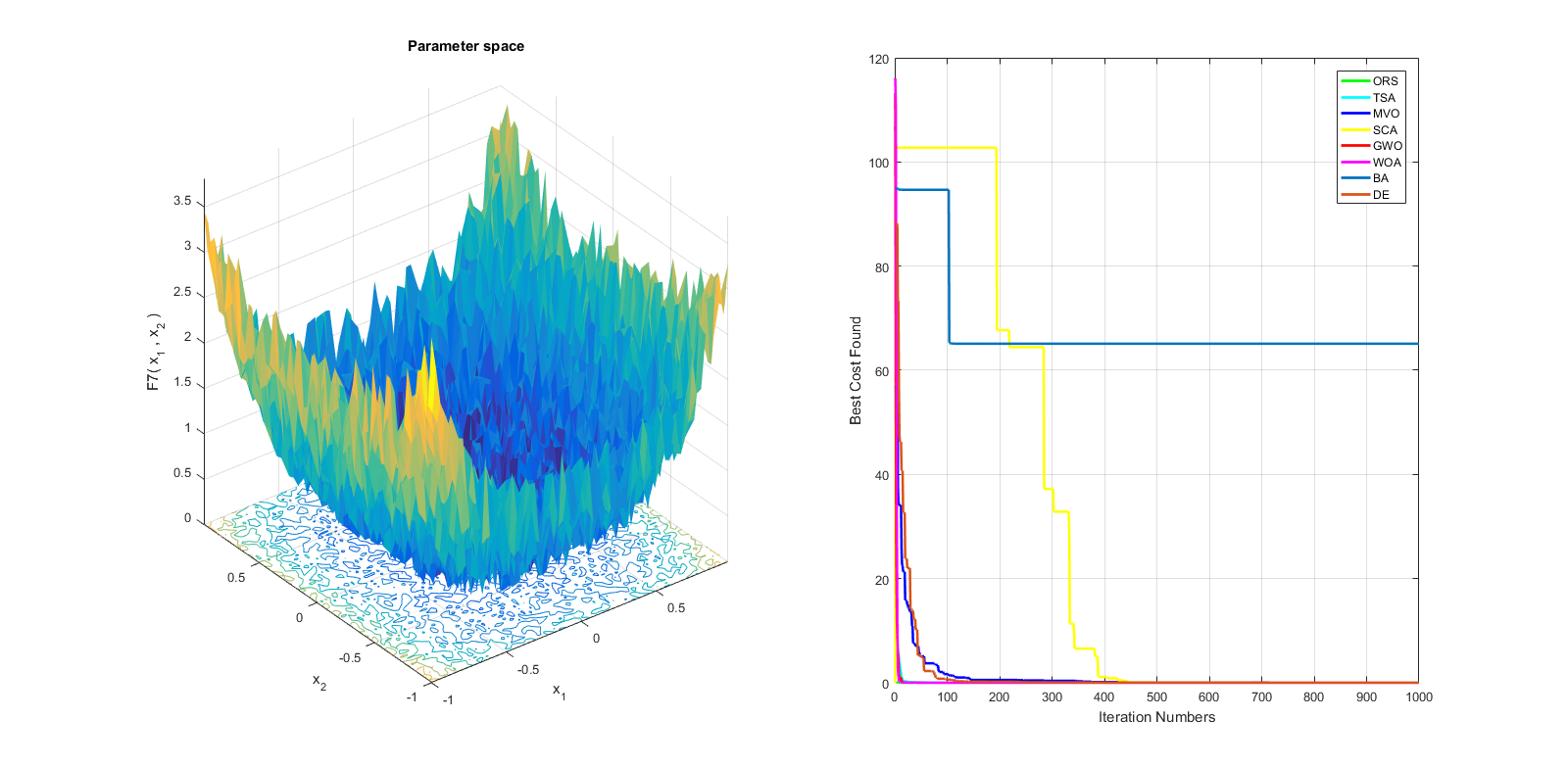}
    \caption{Representation of $Fn_7$ Benchmark function.}
    \label{fig:galaxy}
\end{figure}
\begin{figure}[h]
    \centering
    \includegraphics[width=16cm]{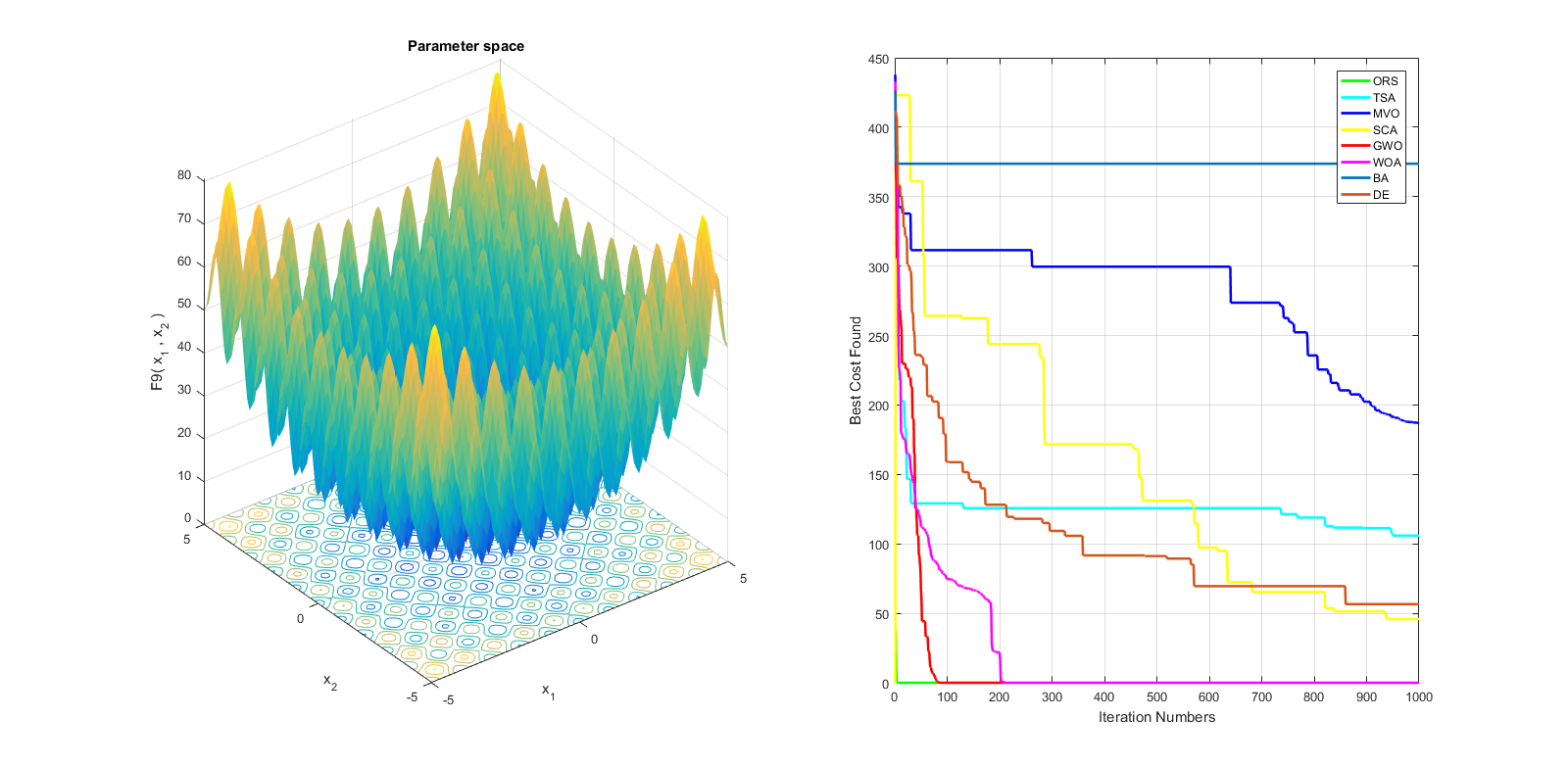}
    \caption{Representation of $Fn_8$ Benchmark function.}
    \label{fig:galaxy}
\end{figure}
\begin{figure}[h]
    \centering
    \includegraphics[width=16cm]{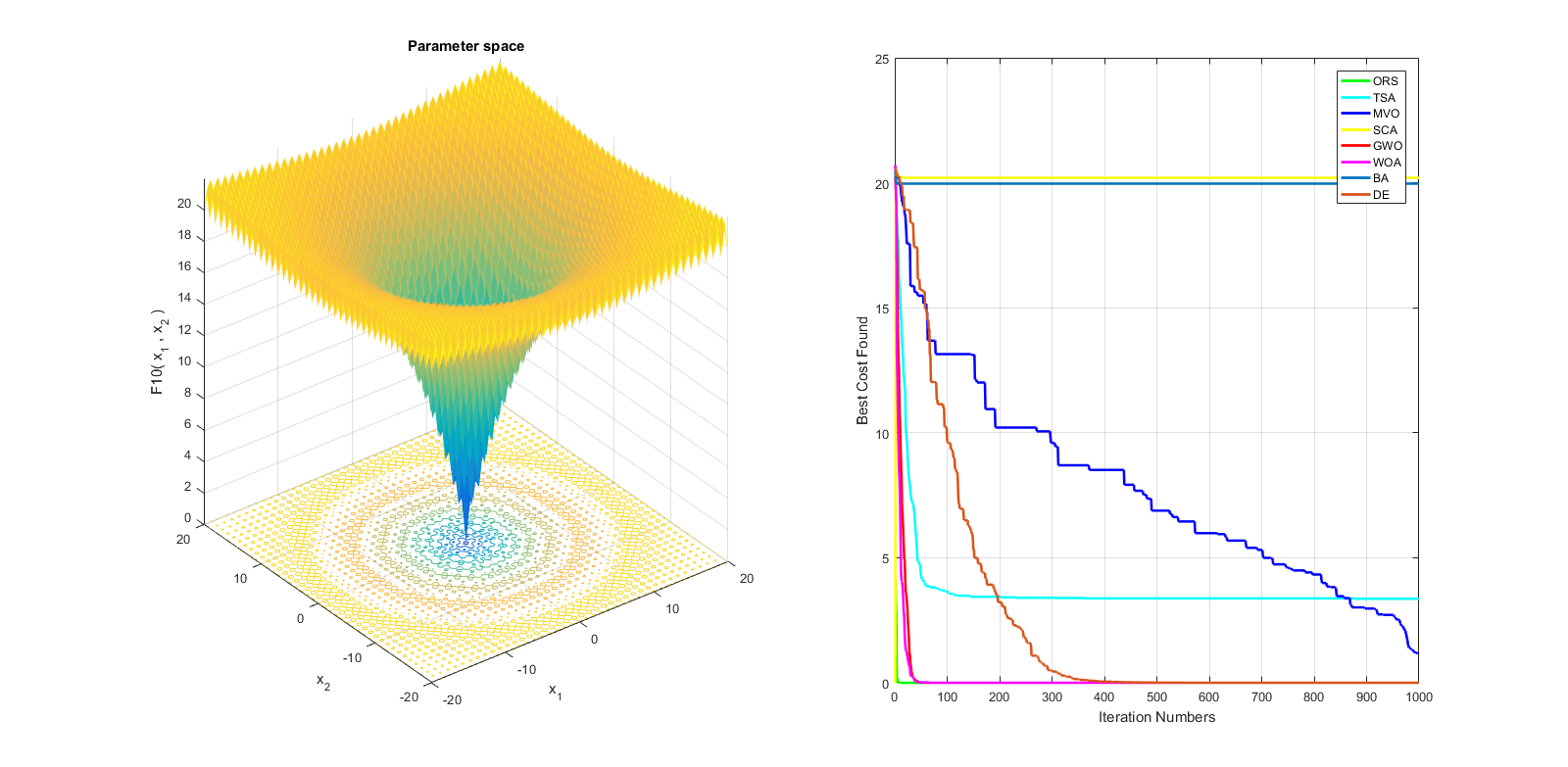}
    \caption{Representation of $Fn_9$ Benchmark function.}
    \label{fig:galaxy}
\end{figure}
\begin{figure}[h]
    \centering
    \includegraphics[width=16cm]{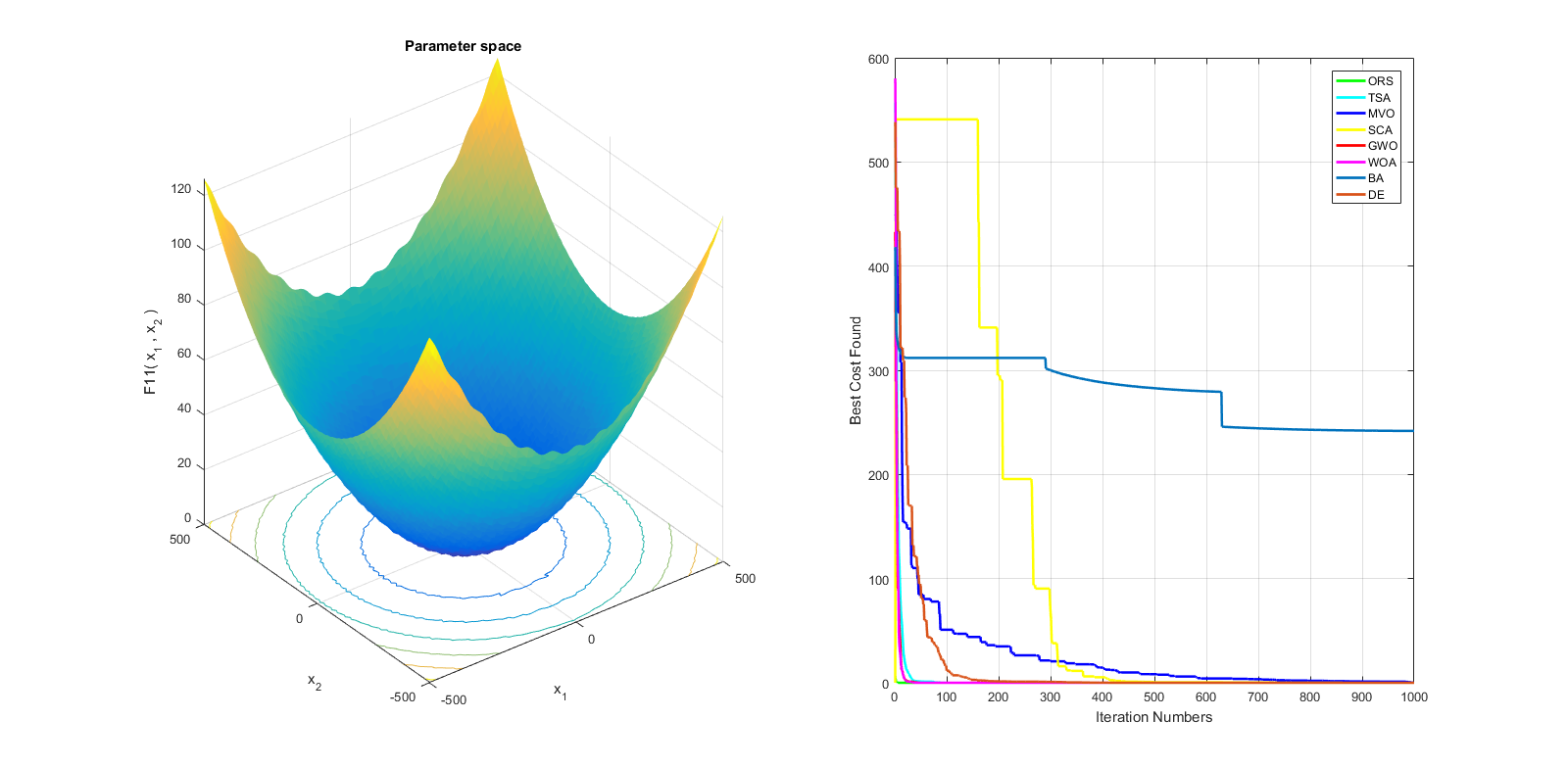}
    \caption{Representation of $Fn_{10}$ Benchmark function.}
    \label{fig:galaxy}
\end{figure}
\begin{figure}[h]
    \centering
    \includegraphics[width=16cm]{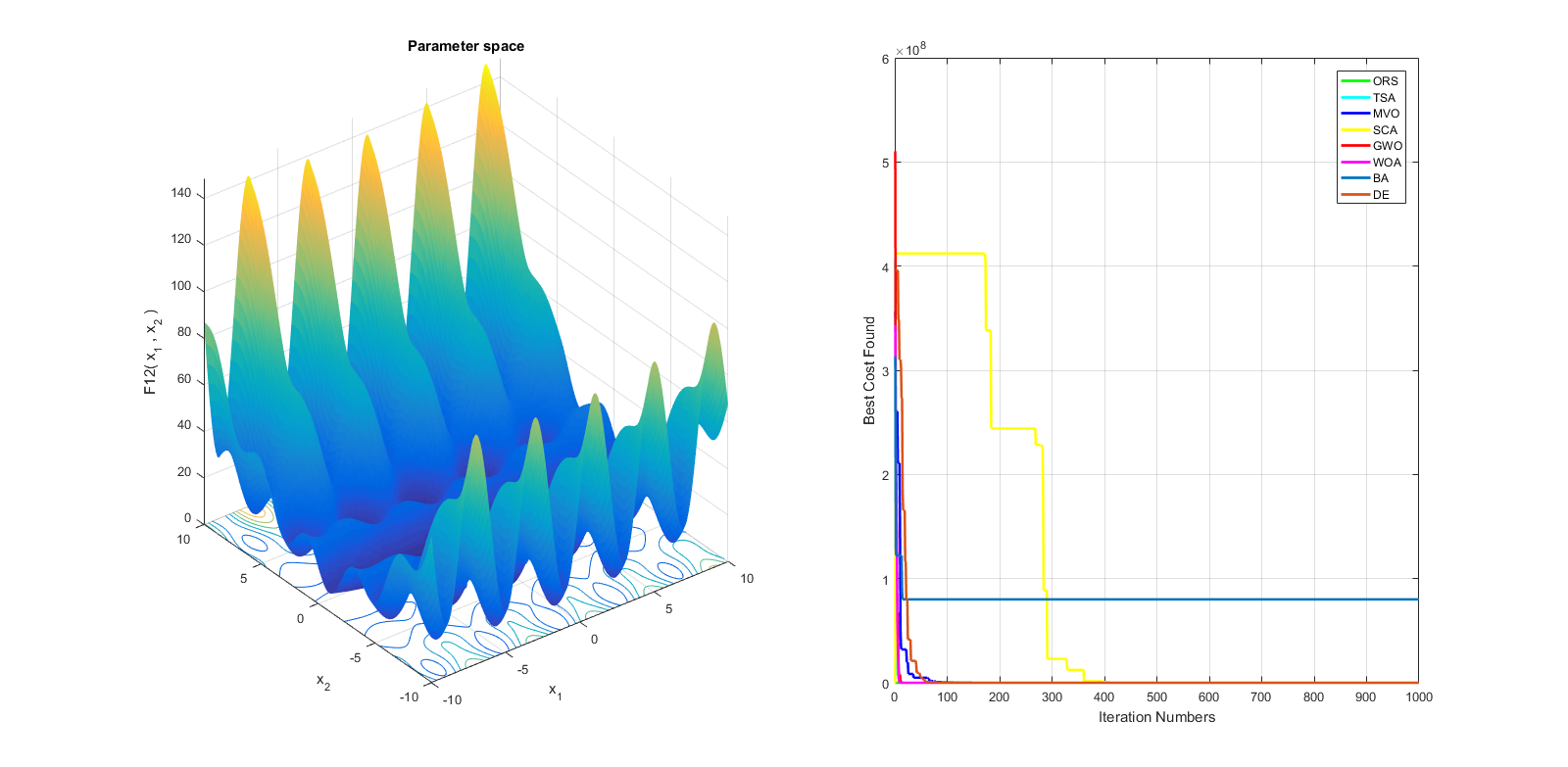}
    \caption{Representation of $F_{11}$ Benchmark function.}
    \label{fig:galaxy}
\end{figure}
\begin{figure}[h]
    \centering
    \includegraphics[width=16cm]{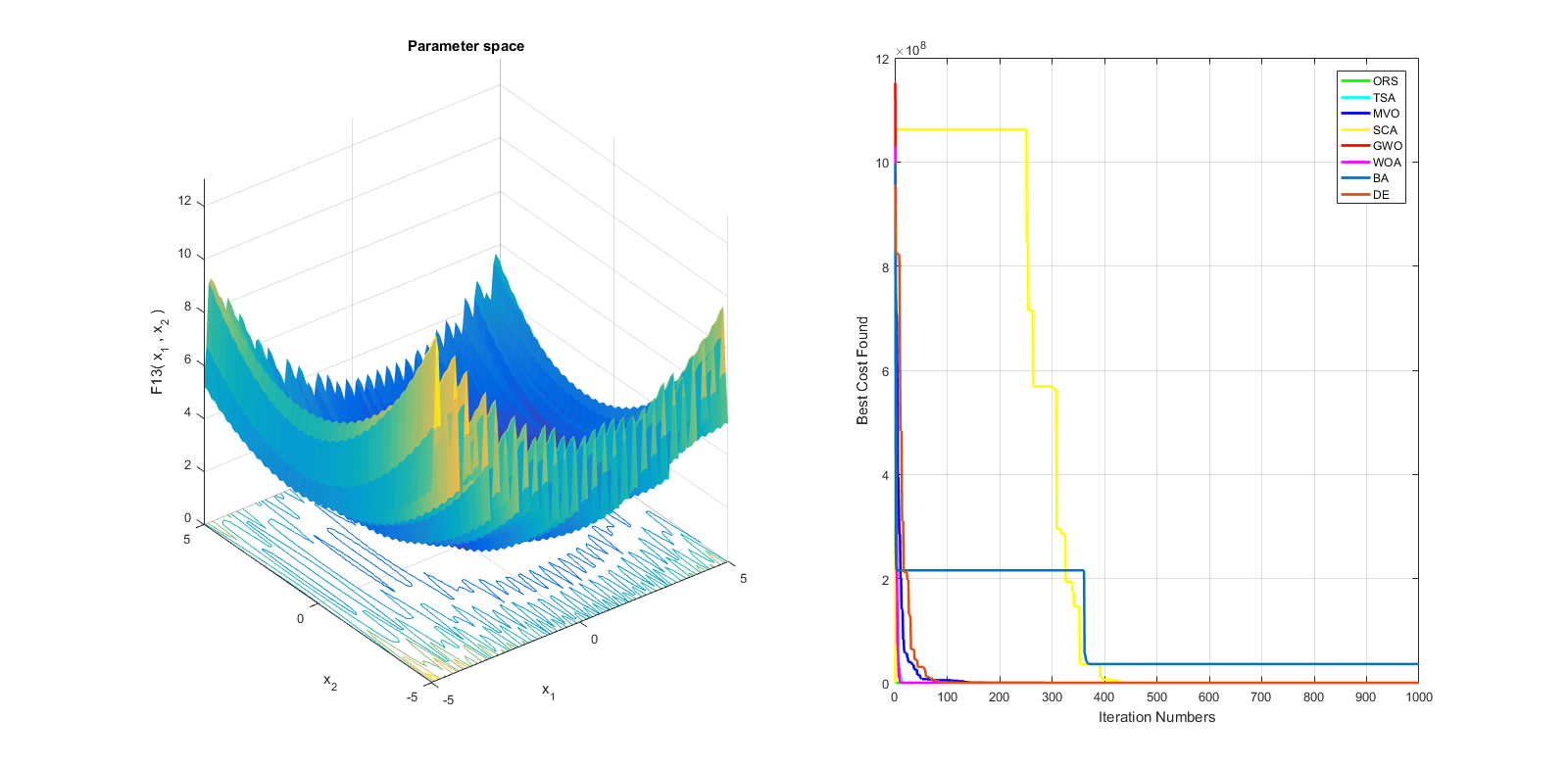}
    \caption{Representation of $F_{12}$ Benchmark function.}
    \label{fig:galaxy}
\end{figure}
\begin{figure}[h]
    \centering
    \includegraphics[width=16cm]{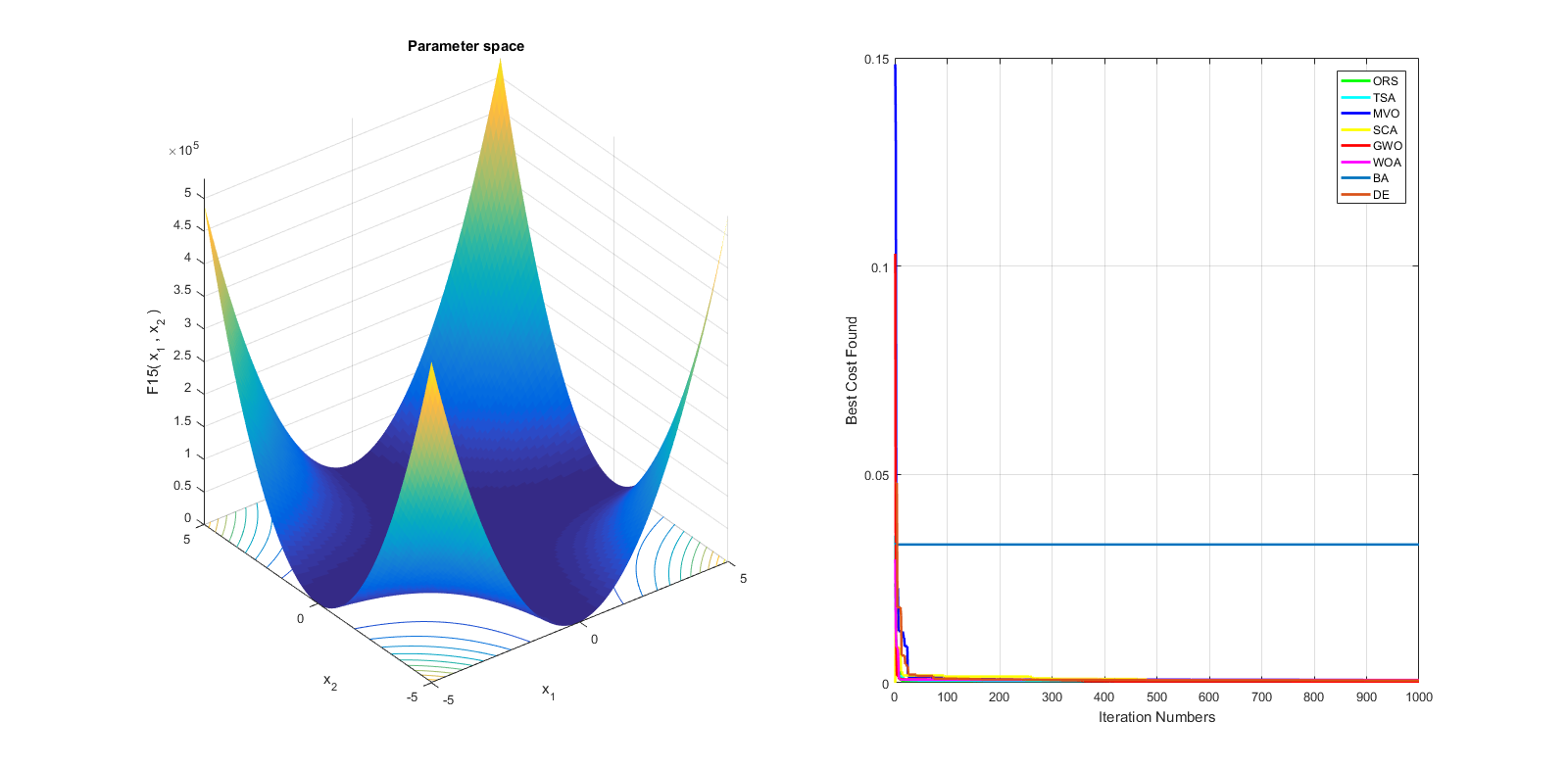}
    \caption{Representation of $F_{13}$ Benchmark function.}
    \label{fig:galaxy}
\end{figure}
\begin{figure}[h]
    \centering
    \includegraphics[width=16cm]{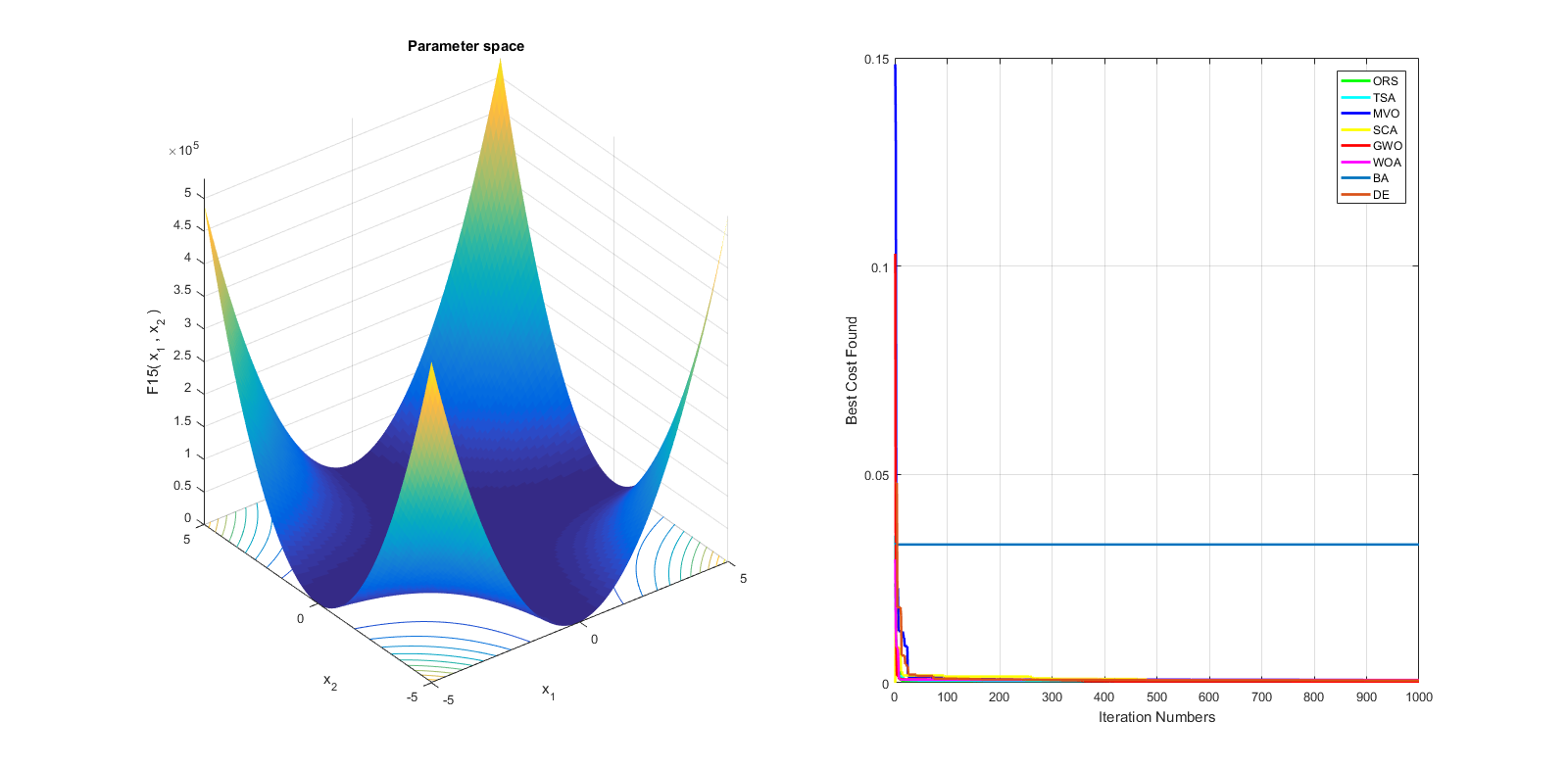}
    \caption{Representation of $F_{14}$ Benchmark function.}
    \label{fig:galaxy}
\end{figure}
\begin{landscape}
\begin{table}[]
\caption{Comparison of results using CEC-06 2019 with 10 Benchmark Functions}
\scriptsize
\begin{tabular}{p{1.2cm}p{1cm}llllllll}
\hline
                   CEC-06 2019 benchmark functions    & Metric            & ORS      & TSA      & MVO      & SCA      & GWO      & WOA      & BA       & DE       \\ \hline
\multirow{2}{*}{CEC01} & mean              & \textbf{1.08E+09} & 1.94E+10 & 4.93E+10 & 1.14E+11 & 1.81E+10 & 8.93E+10 & 1.68E+12 & 8.81E+10 \\ 
                       & standard deviation & 6.06E+08 & 1.25E+10 & 1.49E+10 & 2.25E+10 & 7.11E+09 & 5.86E+10 & 6.98E+11 & 1.81E+10 \\ 
\multirow{2}{*}{CEC02} & mean              & \textbf{31.01788} & 132.2242 & 493.8825 & 210.8078 & 77.976   & 115.3228 & 10702.3  & 307.7471 \\ 
                       & standard deviation & 2.643312 & 41.31204 & 144.9559 & 23.80588 & 15.88247 & 29.78256 & 2730.387 & 23.26109 \\ 
\multirow{2}{*}{CEC03} & mean              & 12.70405 & 12.70251 & 12.70248 & 12.67736 & 12.70244 & 12.70245 & 12.70633 & 12.70253 \\ 
                       & standard deviation & 0.00112  & 6.34E-05 & 2.15E-05 & 0.000206 & 8.82E-06 & 2.36E-05 & 0.000865 & 4.14E-05 \\ 
\multirow{2}{*}{CEC04} & mean              & 23841.01 & 1722.067 & 529.509  & 3723.466 & 470.5797 & 405.2722 & 15324.64 & 529.5947 \\ 
                       & standard deviation & 6439.165 & 1296.329 & 82.91446 & 523.0174 & 39.19761 & 100.2699 & 7717.071 & 151.3487 \\ 
\multirow{2}{*}{CEC05} & mean              & 5.612496 & 2.745676 & 1.90464  & 2.691071 & 1.867342 & 1.70473  & 5.148727 & 1.474339 \\ 
                       & standard deviation & 0.51832  & 0.580735 & 0.060811 & 0.058588 & 0.019892 & 0.252138 & 1.260548 & 0.021004 \\ 
\multirow{2}{*}{CEC06} & mean              & 13.18854 & 11.47623 & 11.36834 & 11.53172 & 11.38292 & 9.897794 & 13.06548 & 9.041339 \\ 
                       & standard deviation & 0.519902 & 0.441242 & 0.440436 & 0.367877 & 0.302417 & 0.957115 & 1.038244 & 0.32729  \\ 
\multirow{2}{*}{CEC07} & mean              & 1661.297 & 619.3856 & 786.3749 & 907.7834 & 783.7534 & 633.9579 & 1291.551 & 416.7321 \\ 
                       & standard deviation & 269.909  & 131.4563 & 136.5048 & 141.2285 & 185.2797 & 252.313  & 440.567  & 19.85074 \\ 
\multirow{2}{*}{CEC08} & mean              & 7.269699 & 6.493685 & 5.762928 & 6.298319 & 4.849311 & 5.647748 & 7.867769 & 5.92149  \\ 
                       & standard deviation & 0.31762  & 0.209493 & 0.75193  & 0.249006 & 0.976601 & 0.615966 & 0.386039 & 0.423799 \\ 
\multirow{2}{*}{CEC09} & mean              & 3603.17  & 188.0933 & 107.3718 & 740.085  & 43.95766 & 82.63873 & 3337.449 & 88.75054 \\ 
                       & standard deviation & 726.7507 & 191.7137 & 11.11833 & 82.92979 & 7.129896 & 19.71339 & 589.9212 & 18.06072 \\ 
\multirow{2}{*}{CEC10} & mean              & 20.79375 & 20.54921 & 20.48398 & 20.48556 & 20.49772 & 20.21186 & 20.90193 & 20.283   \\ 
                       & standard deviation & 0.13009  & 0.072232 & 0.103817 & 0.078978 & 0.094811 & 0.13543  & 0.189459 & 0.026921 \\ \hline
\end{tabular}
\end{table}
\end{landscape}

\begin{figure}
  \centering
   \subfigure(a){\includegraphics[width=1\textwidth]{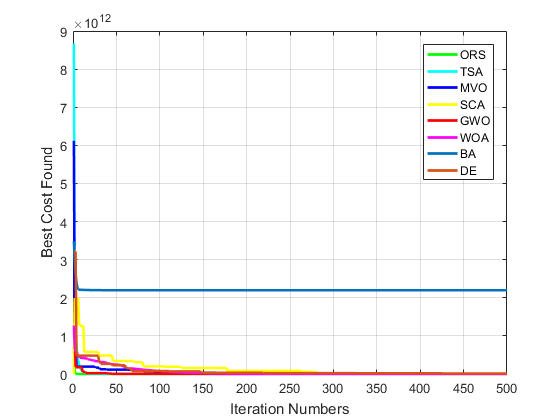}} 
  \subfigure(b){\includegraphics[width=1\textwidth]{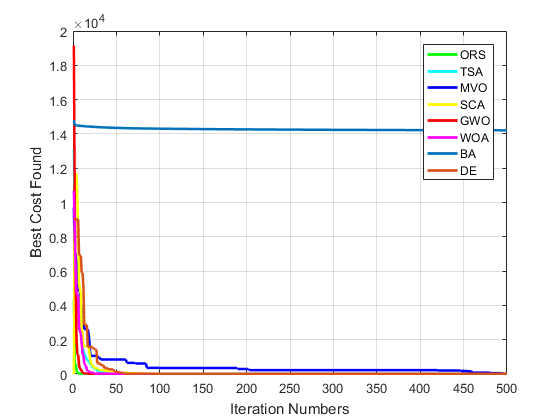}} 
    \caption{(a) Convergence behaviour of CEC01, (b) Convergence behaviour of CEC02 }
 \label{fig:CEC01_02}
\end{figure}

\begin{figure}
  \centering
   \subfigure(a){\includegraphics[width=1\textwidth]{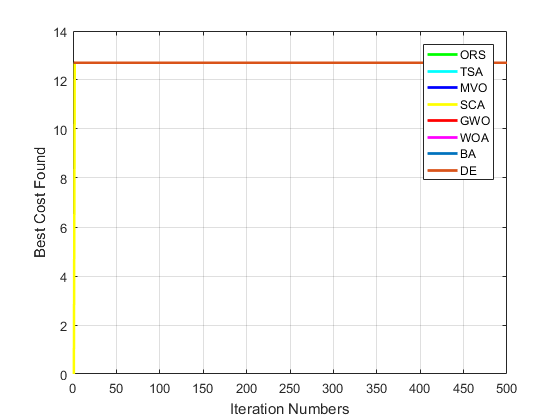}} 
  \subfigure(b){\includegraphics[width=1\textwidth]{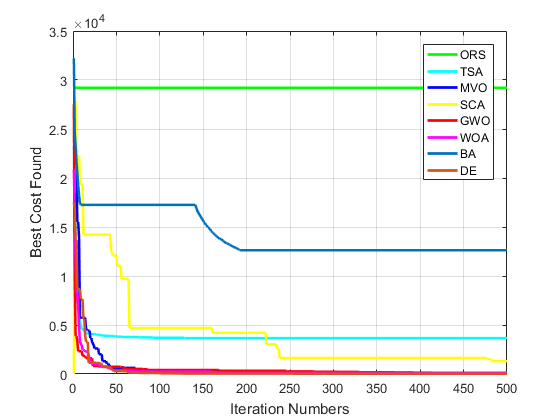}} 
    \caption{(a) Convergence behaviour of CEC03, (b) Convergence behaviour of CEC04 }
 \label{fig:CEC03_04}
\end{figure}

\begin{figure}
  \centering
   \subfigure(a){\includegraphics[width=1\textwidth]{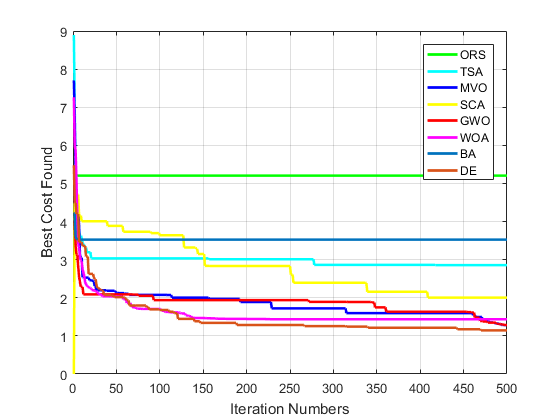}} 
  \subfigure(b){\includegraphics[width=1\textwidth]{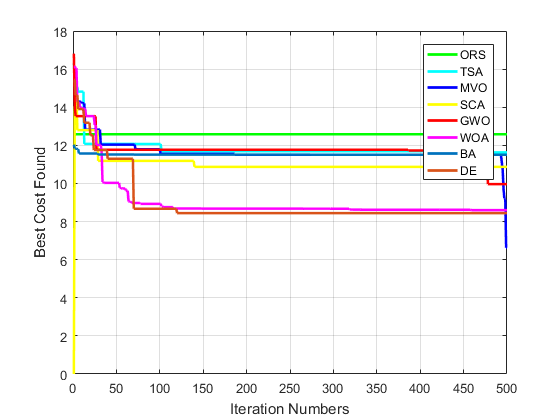}} 
    \caption{(a) Convergence behaviour of CEC05, (b) Convergence behaviour of CEC06 }
 \label{fig:CEC05_06}
\end{figure}

\begin{figure}
  \centering
   \subfigure(a){\includegraphics[width=1\textwidth]{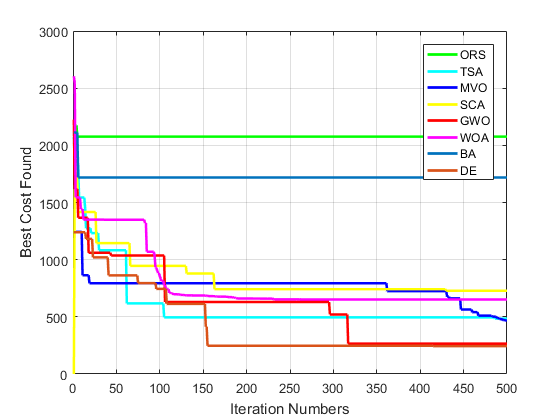}} 
  \subfigure(b){\includegraphics[width=1\textwidth]{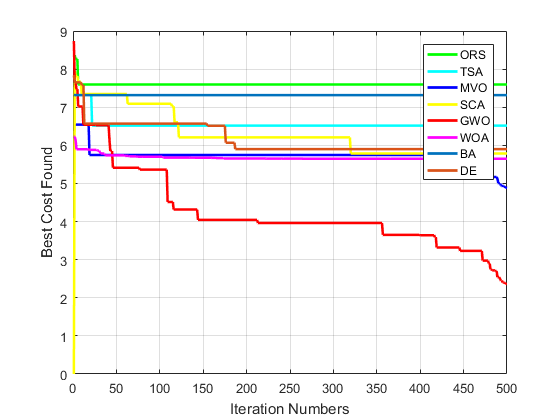}} 
    \caption{(a) Convergence behaviour of CEC07, (b) Convergence behaviour of CEC08 }
 \label{fig:CEC07_08}
\end{figure}

\begin{figure}
  \centering
   \subfigure(a){\includegraphics[width=1\textwidth]{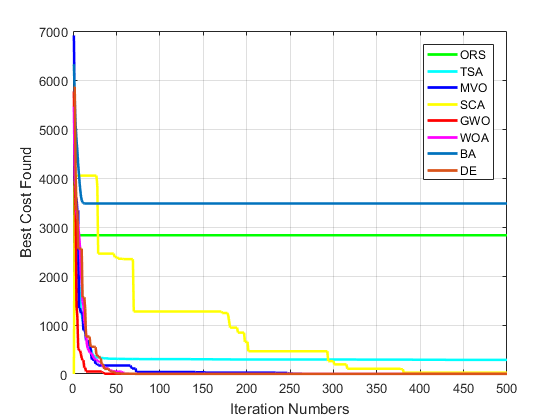}} 
  \subfigure(b){\includegraphics[width=1\textwidth]{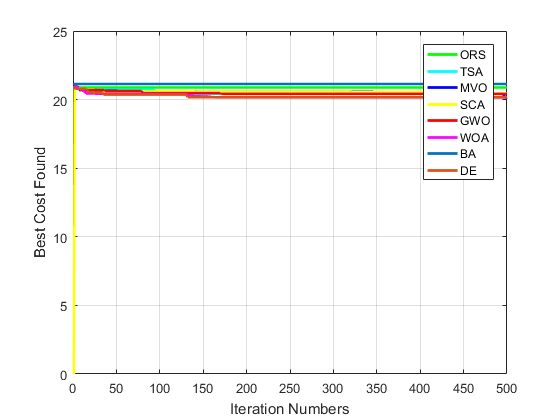}} 
    \caption{(a) Convergence behaviour of CEC09, (b) Convergence behaviour of CEC10 }
 \label{fig:CEC09_10}
\end{figure}

\subsection{Local Minima Avoidance}
ORS shows local minima avoidance in almost all cases by avoiding the multiple local optima and exploiting the global optimum by exploration well as studied from the results in Table 3 and Table 4 of Mean and Standard Deviation. Also, from above analysis of exploitation and exploration it balances well between the exploitation of optimum value by exploring and avoiding the local optima. ORS results are more superior or at par with other meta-heuristics approaches except few cases of CEC-06 2019 test suite CEC04-CEC09 as shown in Figure 25-30.    
\subsection{Convergence Behavior Analysis}
From convergence behavior graphs shown in Figures 8-21, it is observed that the ORS when run on these benchmark functions $Fn_{1}$-$Fn_{14}$ the result reaches the optimum value when the number of iterations increases and it becomes stable when it reaches the minimum optimum value. From the convergence results in all figures represented, it is found that ORS performs well in reaching the $Fn_{min}$ except pre-mature convergence in some cases of CEC-06 2019 test suite \(CEC04-CEC09\) as shown in Figure 25-30 due to falling into local optima. The summary of mean and standard deviation of 10 different runs for CEC-06 2019 benchmark functions are highlighted in Table 5.

\subsection{Statistical Analysis}
The benchmark functions are tested using a standard statistical test known as Wilcoxon signed-rank test for the proposed ORS algorithm to evaluate the mean variation as compared to other algorithms. The results of the Wilcoxon signed-rank test test for all benchmark functions is shown in Table 6 and 7. From the result, it is observed that the value of $p$ in most of the cases is nearer to zero that shows the means of the groups are varying and the null hypothesis is rejected. Only CEC03 shows non-optimal behavior in CEC-06 2019 benchmark functions.

\begin{landscape}
\begin{table}[]
\caption{Wilcoxon signed-rank test test results for benchmark functions}
\begin{tabular}{p{0.7cm}p{0.7cm}p{0.7cm}p{0.7cm}p{0.7cm}p{0.7cm}p{0.7cm}p{0.7cm}p{0.7cm}p{0.7cm}p{0.7cm}p{0.7cm}p{0.7cm}p{0.7cm}p{0.7cm}}
\hline
        & $Fn_{1}$       & $Fn_{2}$       & $Fn_{3}$      & $Fn_{4}$       & $Fn_{5}$       & $Fn_{6}$     & $Fn_{7}$       & $Fn_{8}$       & $Fn_{9}$       & $Fn_{10}$      & $Fn_{11}$      & $Fn_{12}$      & $Fn_{13}$       & $Fn_{14}$      \\ \hline
p-value & 1.89E-63 & 3.54E-64 & 3.60E-40 & 2.77E-33 & 2.10E-26 & 1.98E-51 & 1.46E-11 & 1.81E-82 & 1.25E-39 & 1.92E-41 & 2.25E-29 & 1.85E-88 & 1.19E-06 & 1.78E-63 \\ \hline
\end{tabular}
\vspace{2cm}


\vspace{2cm}
\caption{Wilcoxon signed-rank test test results for CEC-06 2019 benchmark fucntions}

\begin{tabular}{lllllllllllllllllllll}
\hline
Benchmark Functions & \multicolumn{2}{l}{CEC01}     & \multicolumn{2}{l}{CEC02}     & \multicolumn{2}{l}{CEC03}  & \multicolumn{2}{l}{CEC04}  & \multicolumn{2}{l}{CEC05}   & \multicolumn{2}{l}{CEC06}    & \multicolumn{2}{l}{CEC07}      & \multicolumn{2}{l}{CEC08}    & \multicolumn{2}{l}{CEC09}    & \multicolumn{2}{l}{CEC10}     \\ \hline
p-value                                                                    & \multicolumn{2}{l}{0.0001134} & \multicolumn{2}{l}{3.16E-234} & \multicolumn{2}{l}{0.3567} & \multicolumn{2}{l}{0.0951} & \multicolumn{2}{l}{0.00776} & \multicolumn{2}{l}{2.19E-25} & \multicolumn{2}{l}{1.02563-05} & \multicolumn{2}{l}{1.34E-01} & \multicolumn{2}{l}{3.79E-39} & \multicolumn{2}{l}{6.29E-127} \\ \hline
\end{tabular}
\end{table}
\end{landscape}

\subsection {ORS on Engineering Problems}
The ORS algorithm is tested with three classical engineering problems (EP) like Pressure Vessel Design (PVD), Welded Beam Design (WBD), and Spring Design (SD). As per the literature survey performed \cite{19,9,10,4,11,2,1}, these problems are mostly used to test the proposed meta-heuristic approaches. The explicit constraints considered for all engineering problems are the same as given in \cite{19,9,10,4,11,2,1}. Table 8 presents mean optimum value for 10 different runs and standard deviation for different meta-heuristics. It is observed that ORS shows optimal behavior in solving engineering problems. \par
The list of parameters, their range constraints and obtained optimum values for the respective engineering problems are highlighted in Table 9 for all metaheuristics. It is seen that ORS performs well with optimal values of parameters within the given range constraints and satisfying standard problem-specific explicit constraints. The details of standard explicit constraints and its mathematical representations for different engineering problems can be found in \cite{19,9,10,4,11,2,1}.

\subsubsection{Pressure Vessel Design (PVD)}
The Pressure Vessel Design (PVD) problem involves four design variables where $z_{1}$ denotes the thickness of the shell ($T_{s}$), $z_{2}$ represents the thickness of the head ($T_{h}$), $z_{3}$ is used for inner radius ($R$), and $z_{4}$ represents length of the cylindrical section without considering the head ($L$). This problem is a constraint minimization problem where the objective is to minimize the total cost of material, forming, and welding. The mathematical representation of the problem is shown in Equation \ref{20}.

\begin{equation}
\begin{array}{lr@{}c@{}r@{}l}
\text{minimize } \hspace{0.5cm} f(\vec z) = 0.6224z_{1}z_{3}z_{4} + 1.7781 z_{2}z_{3}^2+3.1661z_{1}^2z_{4}+19.84 z_{1}^2z_{3}, \\
\text{subject to }\\
\hspace{2cm}g_{1}(\vec z)=-z_{1}+0.0193z_{3} \le 0 \\
\hspace{2cm}g_{2}(\vec z)=-z_{3}+0.00954z_{3} \le 0\\
\hspace{2cm}g_{3}(\vec z)=-\pi z_{3}^2z_{4}-\frac{4}{3}\pi z_{3}^3+1296000 \le 0\\
\hspace{2cm}g_{4}(\vec z)=z_{4}- 240 \le 0               
                  
\end{array}
\label{20}
\end{equation}

\subsubsection{Welded Beam Design (WBD)}
The Welded Beam Design (WBD) problem considers four design parameters where $z_{1}$ represents weld thickness ($h$), $z_{2}$ denotes the length of attached part of bar ($l$), $z_{3}$ is used for height of the bar ($t$), and  
$z_{4}$ represents thickness of the bar ($b$). The main constraints are the functions of Shear stress ($\tau$), Bending stress ($\sigma$), End deflection ($\delta$), Buckling load ($P_{c}$), and other side constraints are also considered. Here the objective is to minimize the fabrication cost. The mathematical interpretation of this problem is given in Equation \ref{21}.

\begin{equation}
\begin{array}{lr@{}c@{}r@{}l}
\text{minimize } \hspace{0.5cm} f(\vec z) = 1.10471z_{1}^2z_{2} + 0.04811z_{3}z_{4}(14.0+z_{2}), \\
\text{subject to }\\
\hspace{2cm}g_{1}(\vec z)=\tau(\vec z)- \tau_{max} \le 0 \\
\hspace{2cm}g_{2}(\vec z)=\sigma(\vec z)- \sigma_{max}\le 0\\
\hspace{2cm}g_{3}(\vec z)=\delta (\vec z)- \delta_{max} \le 0\\
\hspace{2cm}g_{4}(\vec z)=z_{1}-z_{4} \le 0   \\            
\hspace{2cm}g_{5}(\vec z)=P- P_{c}(\vec z) \le 0 \\ 

\hspace{2cm}g_{6}(\vec z)= 0.125- z_{1} \le 0   \\
\hspace{2cm}g_{7}(\vec z)= 1.10471z_{1}^2z_{2} + 0.04811z_{3}z_{4}(14.0+z_{2})-5.0 \le 0
\end{array}
\label{21}
\end{equation}

\subsubsection{Spring Design (SD)}
The Spring Design (SD) problem involves three design variables where $z_{1}$ is the wire diameter ($d$), $z_{2}$ is mean coil diameter ($D$), and $z_{3}$ is the number of active coils. Here the objective is to minimize the weight of a compression spring subject to constraints like shear stress, surge frequency, and minimum deflection. The mathematical representation of the problem is placed as Equation \ref{22}.

\begin{equation}
\begin{array}{lr@{}c@{}r@{}l}
\text{minimize } \hspace{0.5cm} f(\vec z) = (z_{3}+2)z_{2}z_{1}^2, \\
\text{subject to }\\
\hspace{2cm}g_{1}(\vec z)= 1-\frac{z_{2}^3z_{3}} {71785z_{1}^4} \le 0 \\

\hspace{2cm}g_{2}(\vec z)= \frac{4z_{2}^2-z_{1}z_{2}}{12566(z_{2}z_{1}^3- z_{1}^4)} + \frac{1}{5108z_{1}^2}\le 0\\

\hspace{2cm}g_{3}(\vec z)= 1- \frac{140.45z_{1}}{z_{2}^2z_{3}}\le 0\\
\hspace{2cm}g_{4}(\vec z)=\frac{z_{1}+z_{2}}{1.5}- 1 \le 0               
                  
\end{array}
\label{22}
\end{equation}

\begin{landscape}
\begin{table}[]

\caption{Result comparison of engineering problems}
\scriptsize
\begin{tabular}{p{1cm}p{1.5cm}p{1.2cm}p{1.2cm}p{1.2cm}p{1.2cm}p{1.2cm}p{1.2cm}p{1.2cm}p{1.2cm}}
\hline
EP                            & Metric        & ORS       & TSA      & MVO      & SCA      & GWO      & WOA      & BA       & DE       \\ \hline
\multirow{2}{*}{PVD}            & mean               & \textbf{6384.926}  & 6464.522 & 7007.709 & 7887.301 & 6248.123 & 7618.724 & 17595.43 & 6401.13  \\  
                                                   & standard deviation & 1.35E+02  & 437.4453 & 452.2047 & 621.069  & 117.924  & 1388.93  & 5352.171 & 181.0478 \\ 
\multirow{2}{*}{WBD}                & mean               & \textbf{1.476776}  & 1.507568 & 1.62E+00 & 1.616402 & 1.7449   & 1.70E+00 & 3.30E+00 & 1.636818 \\  
                                                   & standard deviation & 0.0127219 & 0.023827 & 0.031251 & 0.021065 & 0.031065 & 0.13681  & 1.02631  & 0.04616  \\ 
\multirow{2}{*}{SD} & mean               & \textbf{0.012631}  & 0.012791 & 0.013558 & 0.013438 & 0.012887 & 0.013256 & 0.035467 & 0.021345 \\  
                                                   & standard deviation & 0.0000257 & 2.83E-05 & 2.30E-03 & 0.002271 & 0.001647 & 2.29E-03 & 5.86E-02 & 0.03286  \\ \hline
\end{tabular}
\vspace{2cm}
\caption{Parameters and their range constraints for engineering problems and obtained optimum value}
\scriptsize
\begin{tabular}{p{1.5cm}p{1.5cm}llllllll}
\hline
EP & Parameters, Range   \& Optimum Value & ORS      & TSA      & MVO      & SCA      & GWO      & WOA      & BA       & DE       \\ \hline
\multirow{5}{*}{PVD}   & $T_s$: {[}0,99{]}                    & 0.4247   & 0.6347   & 0.4311   & 0.6347   & 0.4347   & 0.4347   & 0.9347   & 0.5347   \\ 
                       & $T_h$: {[}0,99{]}                    & 0.8031   & 0.8471   & 0.8201   & 1.234    & 0.8231   & 0.8231   & 1.6231   & 0.8431   \\ 
                       & R: {[}10,200{]}                      & 42.2675  & 46.2875  & 42.2756  & 47.2843  & 42.2875  & 42.2875  & 84.2875  & 44.2875  \\ 
                       & L: {[}10,200{]}                      & 176.3267 & 179.3789 & 176.1267 & 181.2587 & 176.3567 & 176.3567 & 192.3567 & 179.3567 \\ 
                       & Optimum Value                        & 6031.927 & 6462.522 & 7007.709 & 7884.301 & 6052.345 & 7011.218 & 17595.43 & 6302.13  \\ 
\multirow{5}{*}{WBD}   & h: {[}0.1,2{]}                       & 0.106432 & 0.107831 & 0.109531 & 0.109434 & 0.20435  & 0.20531  & 0.4643   & 0.10531  \\ 
                       & l: {[}0.1,10{]}                      & 2.9118   & 2.942    & 3.1142   & 3.1102   & 3.4641   & 3.4742   & 5.4342   & 3.1742   \\ 
                       & t: {[}0.1,10{]}                      & 8.0976   & 8.1763   & 8.9764   & 8.8763   & 9.0264   & 9.0364   & 12.9876  & 8.9364   \\ 
                       & b: {[}0.1,2{]}                       & 0.10761  & 0.10671  & 0.10962  & 0.10861  & 0.2058   & 0.2061   & 0.4562   & 0.1061   \\ 
                       & Optimum Value                        & 1.476776 & 1.507568 & 1.62E+00 & 1.616402 & 1.7249   & 1.74E+00 & 3.30E+00 & 1.636818 \\ 
\multirow{4}{*}{SD}    & d: {[}0.05,2{]}                      & 0.05062  & 0.05524  & 0.06521  & 0.05223  & 0.05169  & 0.05424  & 0.1721   & 0.1521   \\ 
                       & D: {[}0.25,1.30{]}                   & 0.35317  & 0.366838 & 0.386234 & 0.366231 & 0.356737 & 0.366738 & 1.26235  & 0.96234  \\ 
                       & N: {[}2,15{]}                        & 11.25882 & 11.2983  & 11.4383  & 11.2943  & 11.28885 & 11.2967  & 13.4139  & 12.4231  \\ 
                       & Optimum Value                        & 0.010931 & 0.012798 & 0.013428 & 0.012438 & 0.011187 & 0.012256 & 0.033267 & 0.021145 \\ \hline
\end{tabular}
\end{table}
\end{landscape}
\section{Conclusion \& Future Scope}
This research presented a novel meta-heuristic, ORS, inspired from survival challenges faced by Olive Ridley sea turtle hatchlings. The survival of hatchlings is mainly affected by environmental factors and movement trajectory of hatchlings. These factors are modeled mathematically and the ORS algorithm is analysed theoretically to show the efficacy of the algorithm. Further, it is validated through simulation of fourteen standard mathematical benchmark functions  from CEC test suites of 2005, 2008 and 2010. Also, the behavior of ORS is observed on recent test suite of CEC-06 2019. The  viability of ORS is tested statistically using Wilcoxon signed-rank test test. Three well-known engineering problems are solved optimally using proposed ORS algorithm. The ORS optimzer has shown efficient performance in many of the cases and in some cases, its performance is at par with recent state-of-the-art meta-heuristic optimization algorithms. For some benchmark functions, the sub-optimal behavior of ORS is also observed. The present work only focuses survival modelling of hatchlings in sand area and can be further extended to model other factors at sea level. Also, ORS can be tested on more benchmark functions to explore its performance. Further, the work can be extended by considering multi-population scheme from multiple nests and implementing it on parallel algorithmic models.

\bibliographystyle{unsrtnat}
\bibliography{main}  

\begin{figure}[h!]
  \centering
   \includegraphics[scale=1.75]{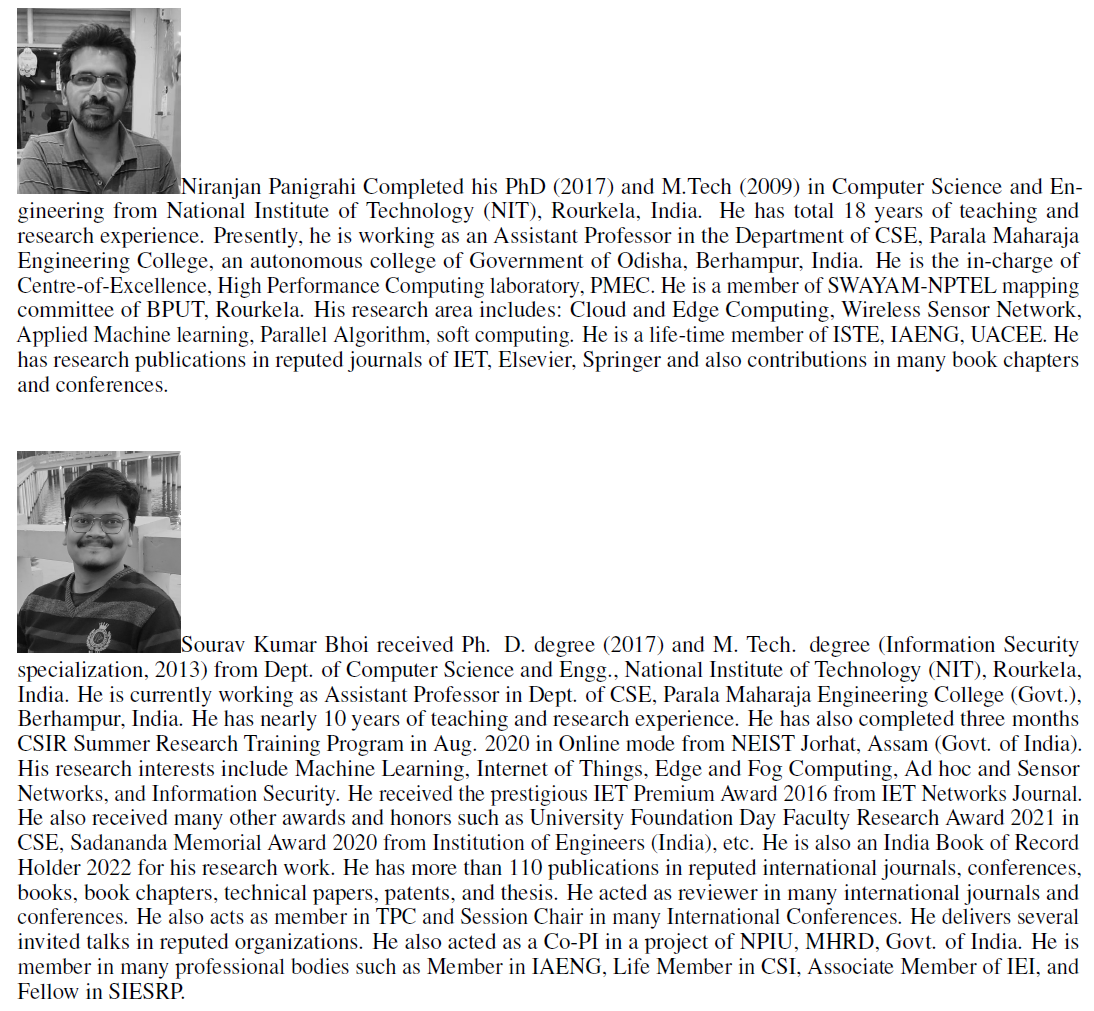}
\end{figure}
\begin{figure}[h!]
  \centering
   \includegraphics[scale=1.75]{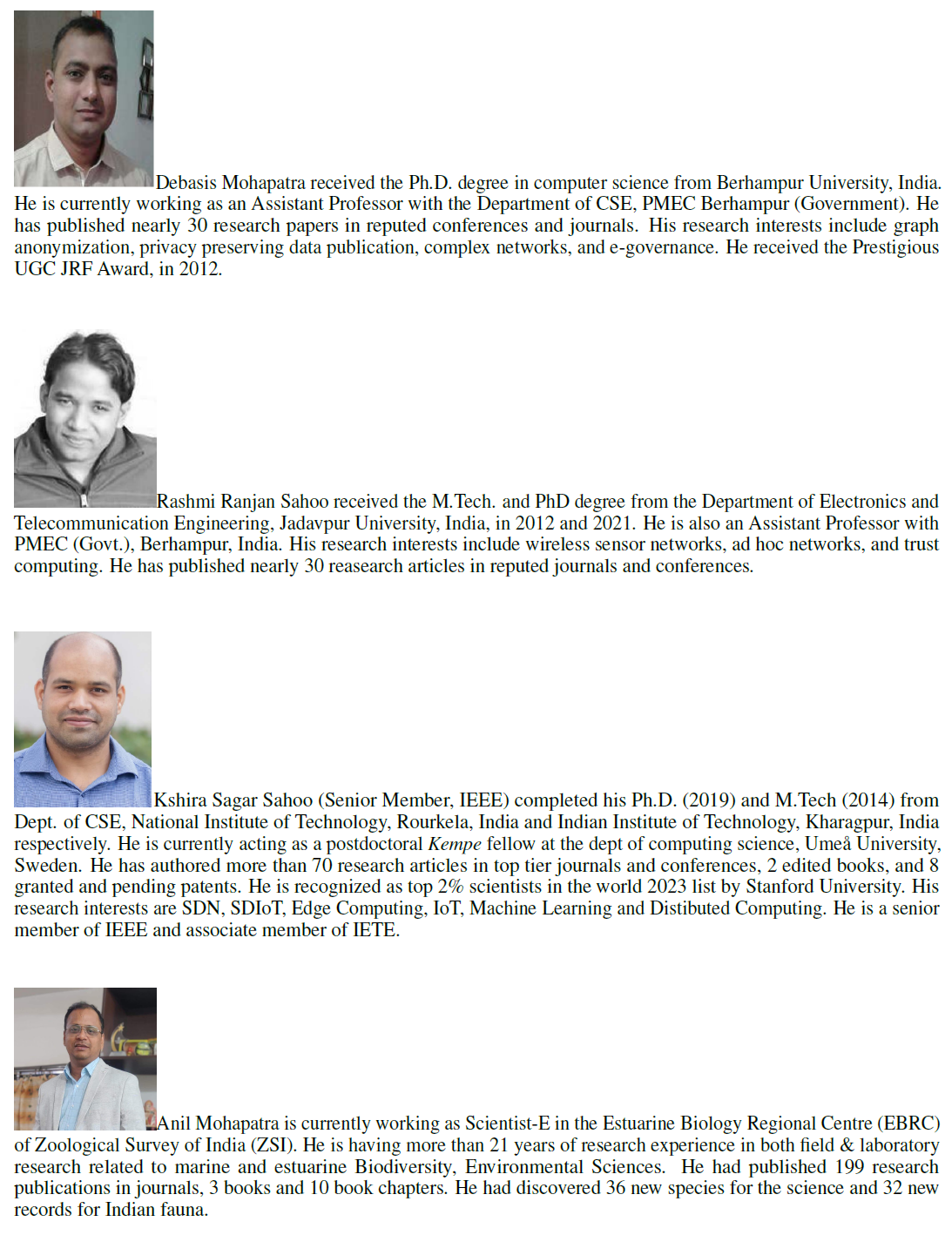}
\end{figure}

\end{document}